\documentclass{article}

\usepackage[final]{neurips_2025}

\usepackage[utf8]{inputenc} 
\usepackage[T1]{fontenc}    
\usepackage{hyperref}       
\usepackage{url}            
\usepackage{booktabs}       
\usepackage{amsfonts}       
\usepackage{graphicx}       
\usepackage{nicefrac}       
\usepackage{microtype}      
\usepackage{multicol}       
\usepackage{xcolor}         
\usepackage{float}
\usepackage{caption}
\usepackage{amsmath}
\usepackage{multirow}
\usepackage{wrapfig}
\usepackage[toc,page]{appendix}
\usepackage{makecell}
\makeatletter
\renewcommand*\env@matrix[1][*\c@MaxMatrixCols c]{%
  \hskip -\arraycolsep
  \let\@ifnextchar\new@ifnextchar
  \array{#1}}
\makeatother

\title{GraLoRA: Granular Low-Rank Adaptation for Parameter-Efficient Fine-Tuning}

\author{
    Yeonjoon Jung$^{1,2*}$ \quad
    Daehyun Ahn$^1$ \quad
    Hyungjun Kim$^1$ \quad
    Taesu Kim$^1$ \quad
    Eunhyeok Park$^{2\dagger}$ \\
    $^1$SqueezeBits \quad
    $^2$POSTECH \\
    \texttt{\{yeonjoon.jung, daehyun.ahn, hyungjun.kim, taesu.kim\}@squeezebits.com} \\
    \texttt{\{yeonjoon.jung, eh.park\}@postech.ac.kr}
}

\begin{document}

\maketitle

\begin{abstract}
Low-Rank Adaptation (LoRA) is a popular method for parameter-efficient fine-tuning (PEFT) of generative models, valued for its simplicity and effectiveness. Despite recent enhancements, LoRA still suffers from a fundamental limitation: overfitting when the bottleneck is widened. It performs best at ranks 32–64, yet its accuracy stagnates or declines at higher ranks, still falling short of full fine-tuning (FFT) performance. We identify the root cause as LoRA’s structural bottleneck, which introduces gradient entanglement to the unrelated input channels and distorts gradient propagation. To address this, we introduce a novel structure, Granular Low-Rank Adaptation (GraLoRA) that partitions weight matrices into sub-blocks, each with its own low-rank adapter. With negligible computational or storage cost, GraLoRA overcomes LoRA’s limitations, effectively increases the representational capacity, and more closely approximates FFT behavior. Experiments on code generation, commonsense reasoning, mathematical reasoning, general language understanding, and image generation benchmarks show that GraLoRA consistently outperforms LoRA and other baselines, achieving up to +8.5\% absolute gain in Pass@1 on HumanEval+. These improvements hold across model sizes and rank settings, making GraLoRA a scalable and robust solution for PEFT.
\end{abstract}

\section{Introduction}
\label{sec:introduction}

Task-specific fine-tuning enables a wide range of applications and significantly improves the quality and effectiveness of generative models. However, the massive scale of these models poses substantial challenges for practical deployment. To address these limitations, Parameter-Efficient Fine-Tuning (PEFT) methods have emerged as a cost-effective alternative~\cite{houlsby2019parameterefficienttransferlearningnlp, xu2023peft_methods}. Among them, Low-Rank Adaptation (LoRA)~\cite{Hu2021LoRA} has gained particular attention for its simplicity and effectiveness, introducing trainable low-rank matrices while keeping the pre-trained model weights frozen. Although the imposed rank-$r$ bottleneck may lead to slight performance degradation compared to full fine-tuning (FFT), its efficiency has led to widespread adoption in practice.

To maximize the benefits of LoRA, various studies have proposed techniques such as improved initialization~\cite{buyukakyuz2024olora,meng2024pissa,paischer2024one,wang2024lora} and structural refinements~\cite{He2025RaSA,Huang2025HiRA,Jiang2024MoRA,Kopiczko2024VeRA} to enhance fine-tuning quality. While these efforts have advanced performance, a substantial quality gap remains compared to FFT, largely due to the inherent upper bound on the rank. Although using a higher rank, within hardware limits, appears to be a natural solution, unfortunately, current implementations of LoRA and its variants do not support such flexibility. Simply increasing the rank often leads to degraded accuracy in many scenarios. 

In this paper, we present a theoretical analysis identifying the root cause of the rank limitation in LoRA. Our analysis reveals a fundamental issue in LoRA’s structure, channel dominance in the gradient, where a small subset of outlier channels disproportionately influences the update direction. This dominance suppresses contributions from other channels, leading to under-utilization of the available rank and degraded performance in tasks that require nuanced or distributed representations.

\begin{figure}
    \centering
    \includegraphics[width=1.0\linewidth]{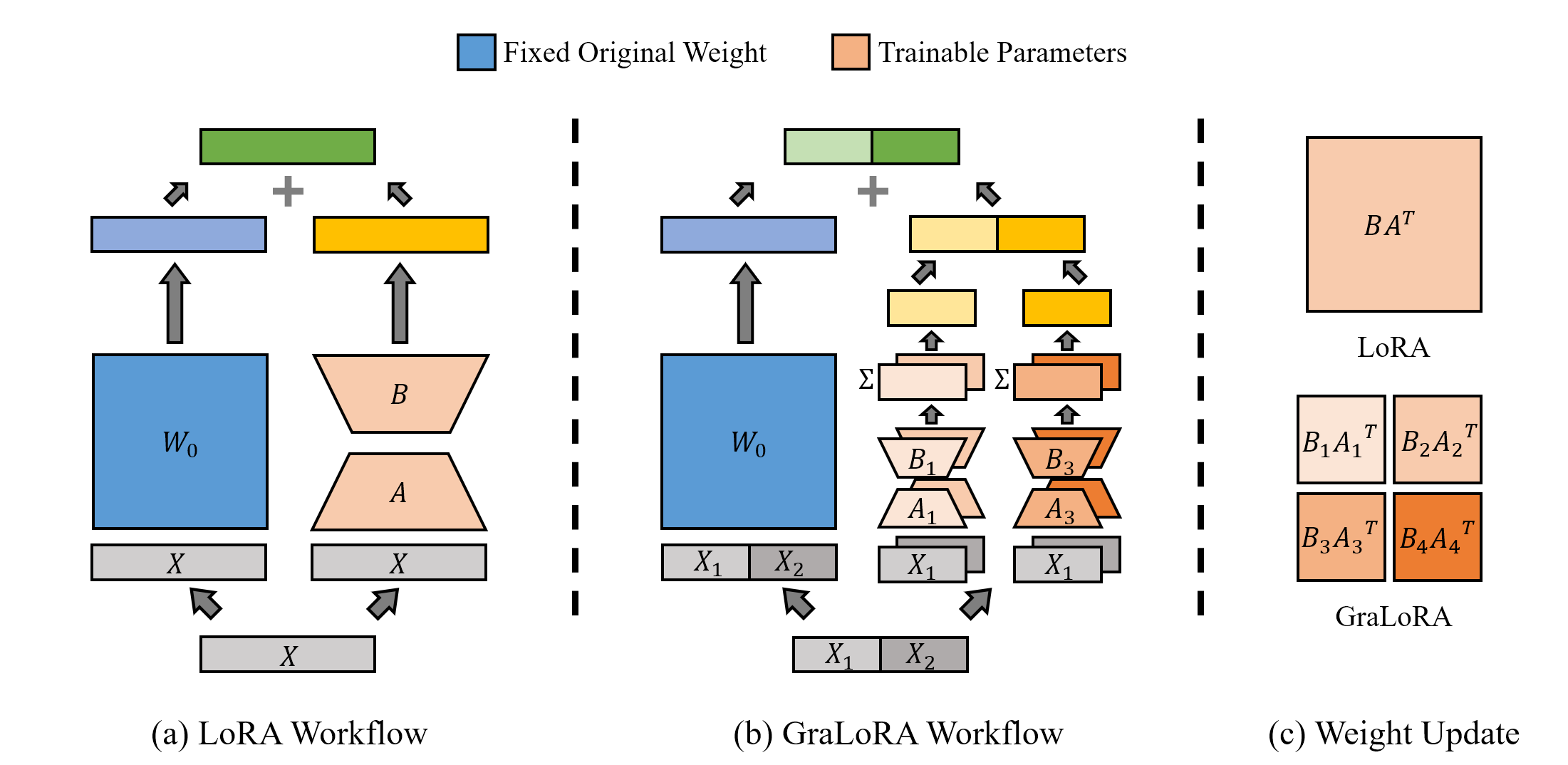}
    \caption{ Illustration of LoRA architecture and GraLoRA architecture. GraLoRA consists of $k^2$ small adapter pairs, where each input and output dimension is $k$ times smaller than the original LoRA.}
    \label{fig:overview}
    \vspace{-0.4cm}
\end{figure}

To overcome these expressivity bottlenecks, we propose Granular Low-Rank Adaptation (GraLoRA), a novel architectural extension of LoRA. As shown in Figure ~\ref{fig:overview}, GraLoRA divides the weight matrix into multiple sub-blocks and applies independent LoRA modules to each, enabling fine-grained updates. This design enhances the model’s capacity to capture complex, localized, or multi-faceted patterns, effectively mitigating the channel dominance issue and improving performance—especially at higher ranks.

Extensive experiments show that GraLoRA consistently outperforms vanilla LoRA across a range of NLP benchmarks, particularly in scenarios with high input heterogeneity or task complexity. These results position GraLoRA as a principled and practical advancement in the PEFT landscape.
~

\section{Details and Limitations of LoRA}
\label{sec:related}

\subsection{Introduction to LoRA}

LoRA is one of the most widely adopted strategies for PEFT. Given a pre-trained weight matrix \( W_0 \in \mathbb{R}^{M \times N} \), where $M$ and $N$ represent the input and output channel dimension, respectively, LoRA keeps $W_0$ frozen and introduces a trainable low-rank update defined as:
\begin{equation}
R = s B A^\top, \quad A \in \mathbb{R}^{N \times r}, \quad B \in \mathbb{R}^{M \times r}, \quad s = \frac{\alpha}{r}.
\end{equation}
Here, rank \( r \) and \( \alpha \) are user-defined hyperparameters. Then, for a given input $X \in \mathbb{R}^{N \times T}$, the output of the LoRA-adapted layer is $Y = W_0 X + R X \in \mathbb{R}^{M \times T}$, where $T$ denotes the batch or token dimension. This low-rank decomposition allows the model to adapt using significantly fewer trainable parameters and reduced memory overhead.

While FFT updates the entire weight matrix, LoRA only updates the decomposed low-rank matrices $A$ and $B$. Note that we assume s = 1 for simplicity, the gradient of the loss with respect to $R$ is:

\begin{equation}
\nabla_{R} L = \frac{\partial L}{\partial R} = \frac{\partial L}{\partial Y} X^\top \in \mathbb{R}^{M \times N}
\end{equation}

From this, the gradients with respect to the LoRA parameters $B$ and $A$ are given by:

\begin{equation}
\frac{\partial L}{\partial B} = \frac{\partial L}{\partial Y} X^\top A, \quad \frac{\partial L}{\partial A^\top} = B^\top \frac{\partial L}{\partial Y} X^\top.
\end{equation}

These result in the following reconstructed update in the fused weight space:

\begin{equation}
\tilde{\nabla}_{R} L = \frac{\partial L}{\partial B} A^\top + B \frac{\partial L}{\partial A^\top} = \frac{\partial L}{\partial Y} X^\top A A^\top + B B^\top \frac{\partial L}{\partial Y} X^\top.
\end{equation}

This expression reveals how the structure of LoRA introduces non-trivial interactions between the gradients and the input, particularly through the rank-$r$ matrices. 

\begin{figure}[t]
    \centering
    \includegraphics[width=1.0\linewidth]{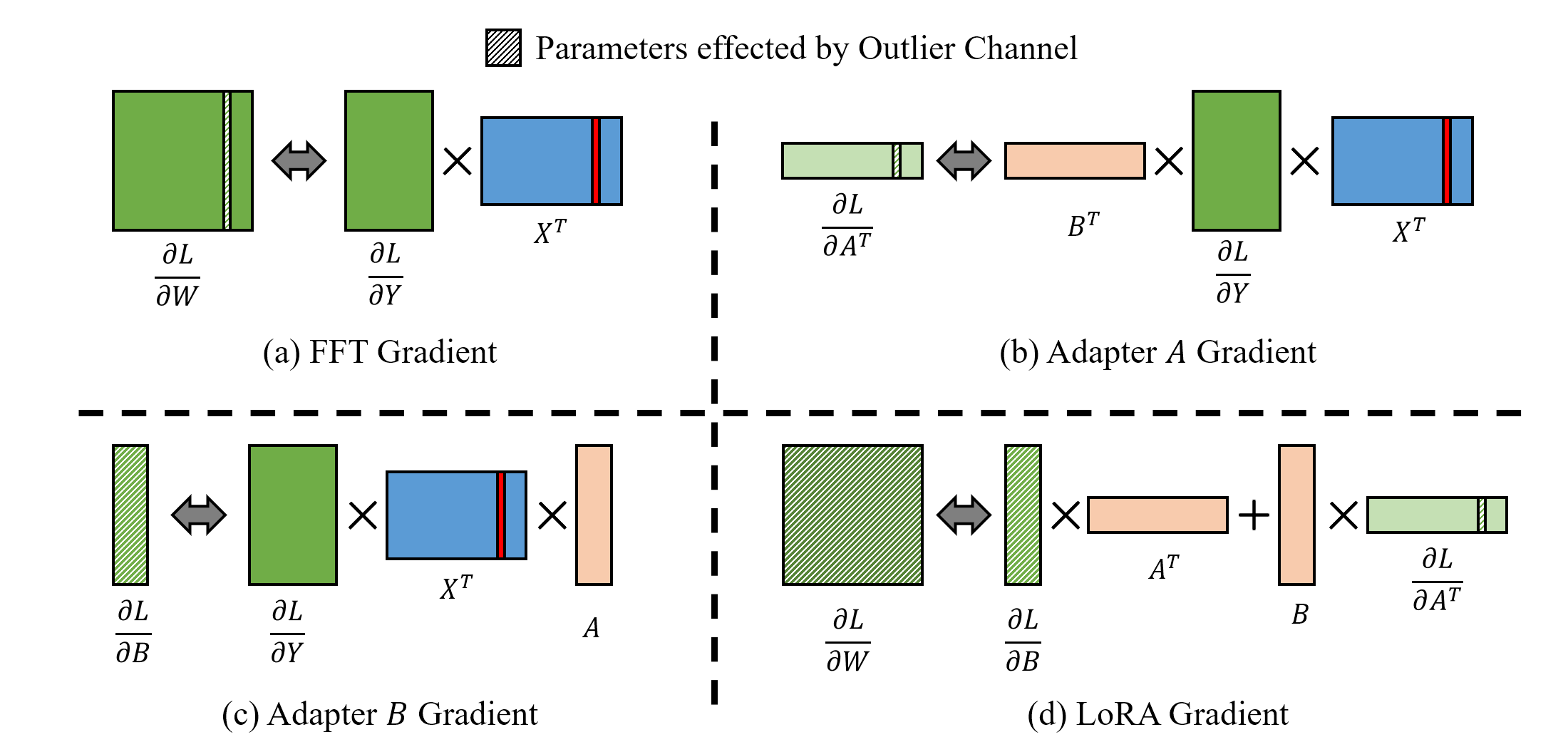}
    \caption{Gradient dynamics of FFT and LoRA in the presence of an outlier input channel. The red channel in input $X$ denotes the outlier. While FFT localizes the gradient impact, LoRA's entire gradient update becomes disproportionately influenced by the single outlier.}
    \label{fig:gradient-update}
    \vspace{-0.4cm}
\end{figure}

\subsection{Why Does LoRA Suffer from a Larger Rank?}

When fine-tuning with a large LoRA rank (e.g., $r > 64$), it is often observed that accuracy degrades compared to using a moderate rank. This counterintuitive behavior arises from the distinct gradient dynamics of LoRA, which differ significantly from those of FFT.

LoRA’s structural design makes its gradients inherently sensitive to the entire input space, as illustrated in Figure~\ref{fig:gradient-update}. In particular, we observe that \textbf{outlier channels}, input channels with abnormally high activations, can disproportionately dominate the gradient signal.

In FFT, the effect of such outliers is typically localized, affecting only a single column of the weight matrix $W$ that directly interacts with the outlier channel. In contrast, LoRA’s low-rank constraint causes the entire gradient of the adapter matrix $B$, denoted $\partial L / \partial B$, to be influenced by these outliers. This results in distorted weight updates in the fused weight space, where the gradient signal from outlier channels overwhelms the contributions from other inputs. Consequently, LoRA fails to accurately replicate the gradient dynamics of FFT, limiting its ability to match FFT-level performance.

\begin{figure}[t]
    \centering
    \includegraphics[width=1.0\linewidth]{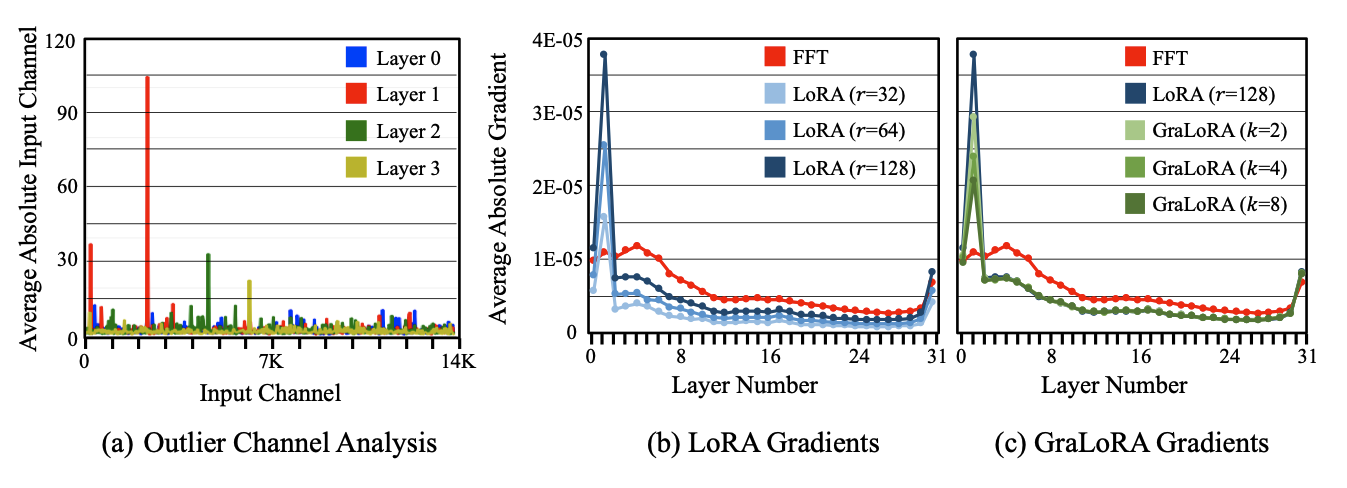}
    \caption{(a) Mean input channel values for the down-projection matrices across layers in LLaMA3.1–8B. A pronounced outlier exists in Layer 1, channel 198 and 2427. (b) Gradient deviation between LoRA and FFT increases with rank, showing LoRA’s susceptibility to input outliers. (c) GraLoRA gradient results at rank 128. GraLoRA noticeably reduces gradient deviation between FFT.}
    \label{fig:outlier_and_layerwise_grads}
\end{figure}

\begin{figure}[t]
    \centering
    \includegraphics[width=1.0\linewidth]{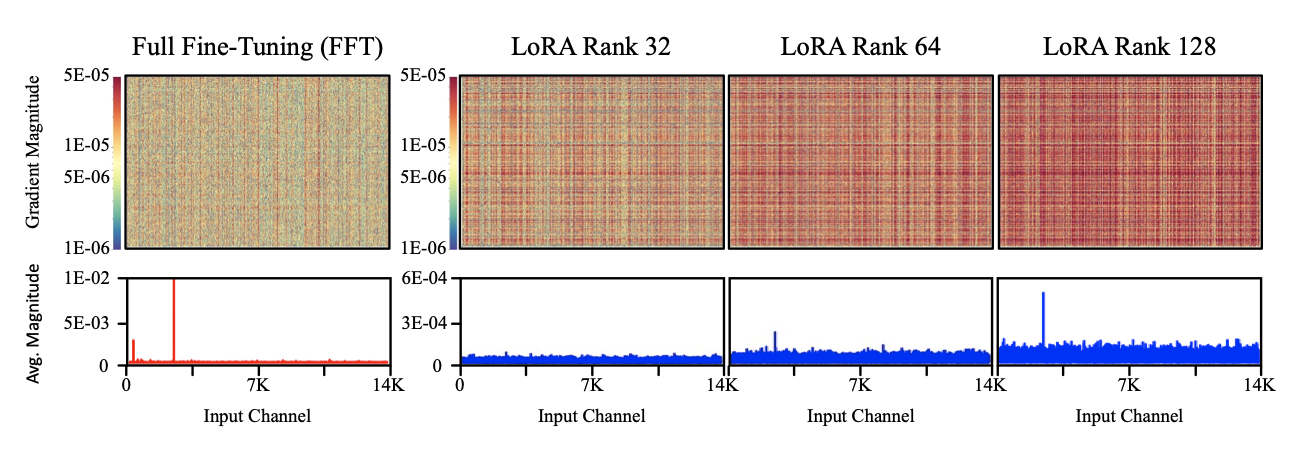}
    \caption{Gradient distribution in Layer 1 down-projection matrix. LoRA gradients show poor alignment with FFT, outlier channel increases the overall gradient scale, while less emphasizing the corresponding outlier channel.}
    \label{fig:fft_lora_grad}
\end{figure}

We observe that in certain layers, most notably the down-projection matrix of Layer 1 in LLaMA3.1–8B, input activations exhibit severe channel-wise imbalance (Figure~\ref{fig:outlier_and_layerwise_grads} (a)). As shown in Figure~\ref{fig:fft_lora_grad}, these outlier channels disproportionately impact the adapter’s gradient updates. 
Figure~\ref{fig:outlier_and_layerwise_grads} further illustrates that the gap between LoRA and FFT gradient updates widens as the LoRA rank increases.

These findings reveal a fundamental misalignment between LoRA updates and the gradient landscape shaped by FFT. The entangled influence of input channels caused by the low-rank projection limits LoRA’s ability to selectively learn from salient features, particularly under skewed input statistics.
While the negative impact of outliers has been well recognized in the context of quantization~\cite{xiao2024smoothquantaccurateefficientposttraining}~\cite{lee2024owq}, their influence on LoRA’s behavior has not been systematically studied until now.

\section{Method}
\label{sec:method}

\subsection{GraLoRA: Granular Low-Rank Adaptation}
\label{sec:gralora}

Motivated by observation in previous section, we propose \textbf{GraLoRA}, a fine-grained and modular extension of LoRA. As illustrated in Figure~\ref{fig:overview}, GraLoRA addresses the limitations of standard LoRA by partitioning the weight matrix into a grid of $k \times k$ independent blocks, each equipped with its own local low-rank adapter. Here, $k$ is a hyperparameter that determines the number of splits along the input and output dimensions. When $k = 1$, GraLoRA reduces to the vanilla LoRA formulation.

Specifically, the weight update \( R \in \mathbb{R}^{M \times N} \) is expressed as the concatenation of block-wise updates:
\begin{equation}\label{eq:gralora-weight}
R_{\text{GraLoRA}} = \begin{bmatrix}
B_{1,1} A_{1,1}^\top & \cdots & B_{1,k} A_{1,k}^\top \\
\vdots & \ddots & \vdots \\
B_{k,1} A_{k,1}^\top & \cdots & B_{k,k} A_{k,k}^\top \\
\end{bmatrix}, \quad A_{i,j} \in \mathbb{R}^{\frac{N}{k} \times \frac{r}{k}}, \quad B_{i,j} \in \mathbb{R}^{\frac{M}{k} \times \frac{r}{k}}
\end{equation}

This block-wise reparameterization provides localized control over each spatial subregion of the parameter space. As detailed in Section~\ref{sec:tradeoff}, GraLoRA incurs the same parameter count and computational overhead as standard LoRA when using the same rank. However, it introduces two key advantages; (1) \textbf{Enhanced Expressivity} and (2) \textbf{Robustness to Input Outliers}. By enabling independent adaptation across $k^2$ subspaces, GraLoRA supports more fine-grained and specialized feature learning. In addition, Localized gradient updates ensure that only the adapters associated with the affected input regions receive large gradients, thereby reducing global gradient distortion and preserving inter-channel signal balance.

\subsection{Expression Power Analysis}
\label{sec:exp-power}

\begin{figure}[t]
    \centering
    \includegraphics[width=1.0\linewidth]{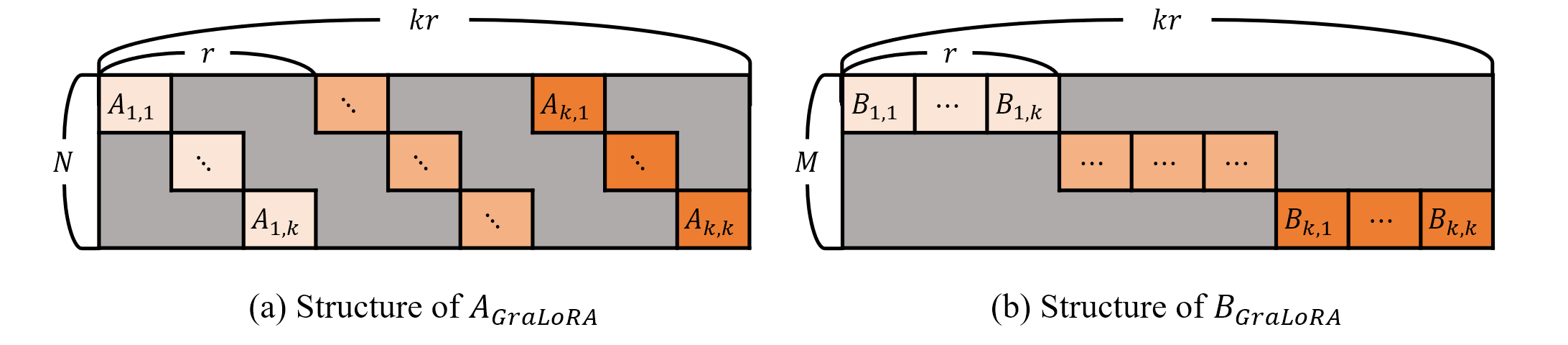}
    \caption{Regularized form of GraLoRA as multiplication of sparse two matrices, \( A_{\text{GraLoRA}} \) and \( B_{\text{GraLoRA}} \).}
    \label{fig:gralora_a_b}
    \vspace{-0.4cm}
\end{figure}
While the weight update of GraLoRA was expressed as concatenation of block-wise updates in (~\ref{eq:gralora-weight}), it can also be regularized as the form of multiplication of two matrices as in the vanilla LoRA. The sparse matrix \( A_{\text{GraLoRA}} \in \mathbb{R}^{ N \times kr } \) can be constructed as Figure ~\ref{fig:gralora_a_b} (a), where \( A_{i, j} \) for \( i, j \in \{ n \in \mathbb{N} \mid n \le k \} \) is located in position \( \left( i + (j-1) \times k, j \right) \) of \( A_{\text{GraLoRA}} \). Other elements are masked out, thus the total number of parameter becomes \( N \times r \).

Then, \( B_{\text{GraLoRA}} \in \mathbb{R}^{ N \times kr } \) is constructed as Figure ~\ref{fig:gralora_a_b} (b), where matrix \( B_{i, j} \) for \( i, j \in \{ n \in \mathbb{N} \mid n \le k \} \) is located in position \( \left( i, j + (i-1) \times k \right) \) of \( B_{\text{GraLoRA}} \), Similarly, other composition of the matrix is masked, therefore the total number of parameter becomes \( M \times r \). Then the weight update of GraLoRA can be expressed as \( W = W_0 + R_{\text{GraLoRA}} = W_0 + B_{\text{GraLoRA}} A_{\text{GraLoRA}}^\top \).

Assuming that all columns of $\left[ B_{i,1}, \cdots, B_{i,k} \right]$ are linearly independent, the rank of $B_{\text{GraLoRA}}$ becomes $\mathbf{R}\left( B_{\text{GraLoRA}} \right) = kr$. Similarly, if all columns of $\left[ A_{1,j}, \cdots, A_{k,j} \right]$ are linearly independent, the rank of $A_{\text{GraLoRA}}$ is $\mathbf{R}\left( A_{\text{GraLoRA}} \right) = kr$.
Applying Sylvester’s rank inequality to derive the lower bound and the matrix product theorem for the upper bound, we obtain:
\begin{equation}
    \mathbf{R}(B_{\text{\tiny{GraLoRA}}}) + \mathbf{R}(A_{\text{\tiny{GraLoRA}}}^\top) - kr \le \mathbf{R}(B_{\text{\tiny{GraLoRA}}} A_{\text{\tiny{GraLoRA}}}^\top) \le min(\mathbf{R}(B_{\text{\tiny{GraLoRA}}}), \mathbf{R}(A_{\text{\tiny{GraLoRA}}}^\top))
\end{equation}
Thus, the effective rank of $R_{\text{GraLoRA}}$ becomes $kr$, which is $k$ times higher than that of the vanilla LoRA method—effectively enhancing the model’s expressive capacity. The rank analysis of fine-tuned LoRA and GraLoRA, summarized in Table~\ref{tab:rank_size} in Appendix, demonstrates that GraLoRA linearly scales the representational power of the adaptation matrix in practical settings.

\subsection{Gradient Dynamics Under Outlier Activation}
\label{sec:gralora-gradients}

GraLoRA effectively localizes the influence of outlier channels to a limited subset of adapter blocks. Because each block processes only a specific slice of the input, only the $k$ adapter pairs intersecting with the outlier channel are exposed to amplified gradients. In contrast, the remaining $k^2 - k$ adapters maintain gradient magnitudes close to baseline levels. This selective gradient propagation resembles the behavior of FFT, where only weights directly connected to active inputs are significantly updated.

\begin{figure}[t]
    \centering
    \includegraphics[width=1.0\linewidth]{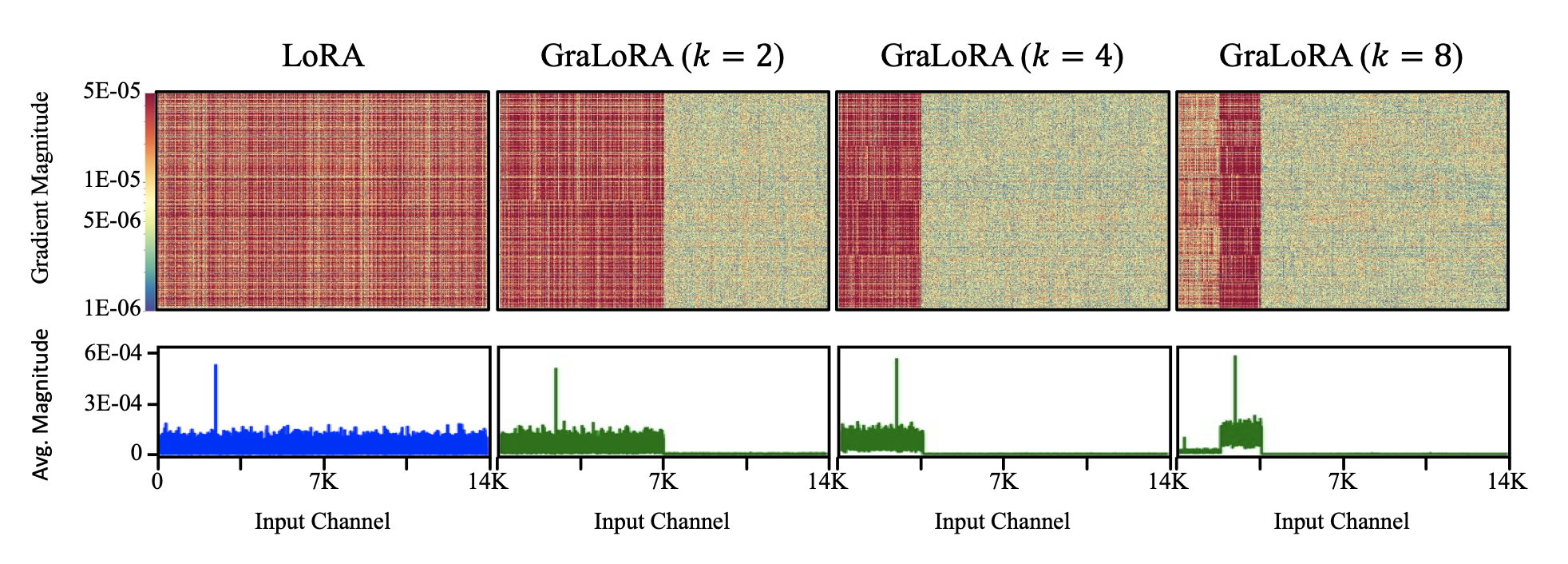}
    \caption{Comparison of gradient distributions under outlier activation. In GraLoRA, only the blocks interacting with the outlier exhibit elevated gradients, mitigating global distortion and aligning with FFT behavior.}
    \label{fig:gralora_grad}
    \vspace{-0.2cm}
\end{figure}

GraLoRA's impact on gradient dynamics can be observed by comparing gradient distributions of the down-projection matrix in Layer 1 with standard LoRA. As illustrated in the Figure~\ref{fig:outlier_and_layerwise_grads} (c) and Figure~\ref{fig:gralora_grad}, GraLoRA reduces the gradient deviation and limits the influence of outlier channels, overcoming the limitations of standard LoRA with larger ranks.

\subsection{Tradeoff Analysis}
\label{sec:tradeoff}

As discussed, GraLoRA provides several advantages over standard LoRA. However, these benefits do not come without cost. In this section, we provide deeper analysis on the overhead introduced by GraLoRA.

\textbf{Computation Overhead Analysis:} First, we analyze the expected computational cost of LoRA in terms of FLOPs. To take advantage of the low-rank structure, LoRA computes the projection in two sequential steps. The first computes \( A^\top X \in \mathbb{R}^{r \times T} \), followed by the reconstruction \( B(A^\top X) \in \mathbb{R}^{M \times T} \). These steps require \( 2NrT \) and \( 2rMT \) FLOPs, respectively, resulting in a total complexity of $
O\left( r(M + N)T \right).$

Similarly, GraLoRA divides the computation into two steps involving $k^2$ adapter blocks. In the first step, the projection computes $A_{i,j}^\top X_j \in \mathbb{R}^{\frac{r}{k} \times T}$ for each of the $k^2$ blocks, incurring a total cost of
$
2 \cdot \frac{N}{k} \cdot \frac{r}{k} \cdot T \cdot k^2 = 2NrT.
$ In the second step, each intermediate output is processed by its corresponding $B_{i,j}$, producing $B_{i,j}(A_{i,j}^\top X_j) \in \mathbb{R}^{\frac{M}{k} \times T}$. This step adds another
$
2 \cdot \frac{r}{k} \cdot \frac{M}{k} \cdot T \cdot k^2 = 2rMT.
$ FLOPs to the total cost. Hence, the overall computational cost of GraLoRA remains $O\left( r(M + N)T \right)$, maintaining efficiency comparable to vanilla LoRA while significantly enhancing expressive power. A detailed analysis of computational overhead is provided in Appendix~\ref{sec:compute}.

\begin{table}[H]
    \centering
    \vspace{-0.3cm}
    \caption{Maximal allocated memory during training LLaMA3.1–8B model with batch size 1. Input length was set to 1024 and memory allocated for weight was removed for direct comparison.}
    \resizebox{0.8\textwidth}{!}{
    \begin{tabular}{c|cccc}
        \toprule
         & LoRA & GraLoRA (k=2) & GraLoRA (k=4) & GraLoRA (k=8) \\
        \midrule
        Vanilla Backward (GB) & 10.0 & 10.1 & 10.2 & 10.4 \\ 
        \midrule
        Gradient Checkpointing (GB) & 2.6 & 2.6 & 2.6 & 2.6 \\ 
        \bottomrule
    \end{tabular}
    }
    \label{tab:memory}
    \vspace{-0.3cm}
\end{table}

\textbf{Memory Overhead Analysis:} As with classical LoRA, GraLoRA can be merged into the original weight matrix at inference time. Therefore, our analysis focuses on the memory overhead incurred during training. Although the number of parameters and FLOPs are identical to those of LoRA, the intermediate latent representation \( A_{\text{GraLoRA}}^\top X \) becomes \( k \) times larger than the corresponding \( A^\top X \) in standard LoRA. This expanded latent space allows for greater information preservation, which can be beneficial. However, it also leads to increased memory consumption during training time. Fortunately, the rank \( r \) is typically much smaller than the input and output dimensions, thus the additional memory required remains marginal—even for large \( k \), as demonstrated in Table~\ref{tab:memory}. Moreover, by applying recent techniques such as gradient checkpointing, the memory overhead from the expanded latent space can be effectively hidden, making the impact negligible in practice.

\textbf{Selection of k}
While GraLoRA increases the total rank from \( r \) to \( kr \), each individual block, represented as \( B_{i,j} A_{i,j}^\top \in \mathbb{R}^{\frac{M}{k} \times \frac{N}{k}} \), is constrained to a reduced rank of \( \frac{r}{k} \). As a result, increasing \( k \) beyond a certain threshold can degrade performance due to limited expressiveness within each block.
This effect is especially pronounced when the overall rank \( r \) is small. Empirically, we observed that maintaining a minimum block expressiveness of approximately \( r/k^2 \approx 8 \) yields stable performance across various configurations. Based on this observation, we adopted \( k = 2 \) for ranks 16 and 32, and \( k = 4 \) for ranks 64 and 128 in our experiments. Detailed \( k \)-sweep results can be found in Section~\ref{sec:ablation}.

\subsection{Hybrid GraLoRA}

\begin{figure}[t]
    \centering
    \includegraphics[width=1.0\linewidth]{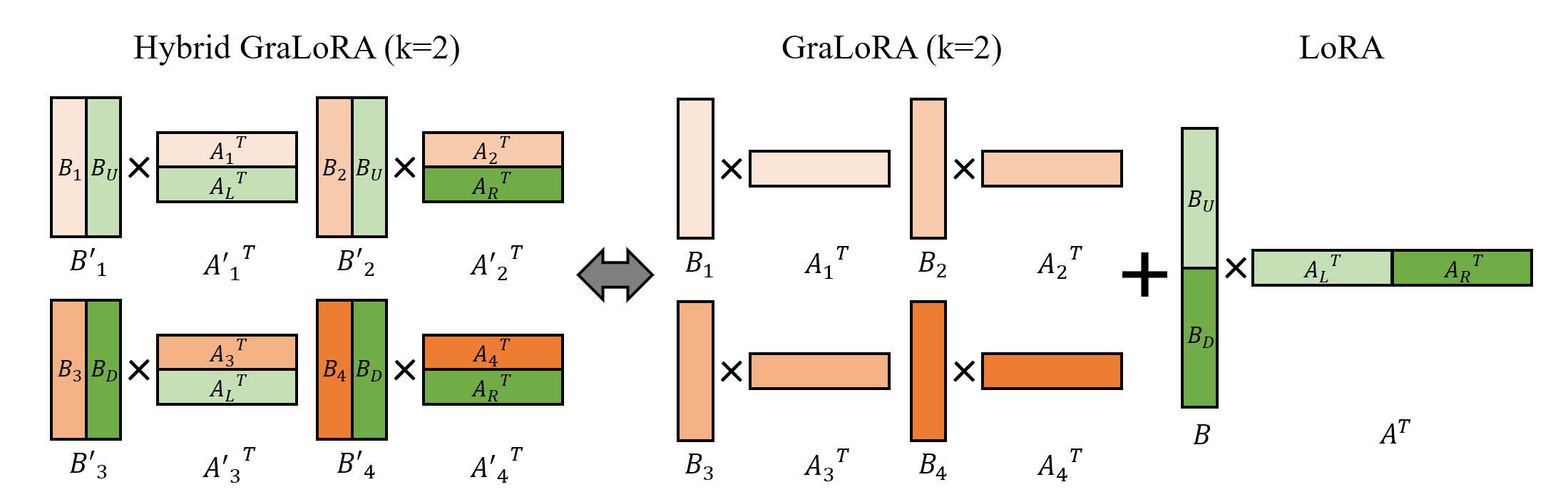}
    \caption{Hybrid GraLoRA architecture when GraLoRA \(k=2\). LoRA parameter becomes shared across small GraLoRA adapters in the same row or same column.}
    \label{fig:hybrid_gralora}
    \vspace{-0.4mm}
\end{figure}

On the other hand, for smaller ranks—typically rank 16 or below—using \( k = 2 \) may still lead to performance degradation or yield only marginal gains. To address this limitation, we introduce a hybrid approach that combines the strengths of LoRA and GraLoRA. This method retains the fine-grained input handling and increased total rank offered by GraLoRA, while preserving the expressive power of larger block units through LoRA. Since LoRA shares the same parameters across both rows and columns, it can be naturally integrated with GraLoRA in a concatenated form, which we refer to as \emph{Hybrid GraLoRA} (see Figure~\ref{fig:hybrid_gralora}). Empirically, we found that allocating up to \( \frac{1}{2} \) of the total rank to the LoRA component mitigated the limitations of GraLoRA in low-rank scenarios ($\gamma <= 16$), while fully allocating the rank to GraLoRA better performed in high-rank circumstances.

\section{Experiments}
\label{sec:experiments}
In order to validate the superiority of the proposed idea, we conduct an extensive analysis on large-scale dataset with the state-of-the art LLMs. 
We evaluate GraLoRA across five challenging domains: \textbf{code generation}, \textbf{commonsense reasoning}, \textbf{mathematical reasoning}, \textbf{general language understanding}, and \textbf{personalized image generation}. Our experiments are designed to assess whether the proposed granular adaptation mechanism improves performance across varying model sizes, LoRA ranks, and tasks that require nuanced reasoning and high representational fidelity.

\subsection{Experimental Setup}

\textbf{Code Generation.} We fine-tuned LLaMA3.1–8B (~\cite{grattafiori2024llama3herdmodels}) with 4 A100 80G GPU on the Magicoder-Evol-Instruct-110k~\cite{wei2024magicoder} train dataset, a curated and decontaminated subset of WizardCoder~\cite{luo2024wizardcoder}, comprising high-quality instruction–response pairs for programming tasks. Evaluation was conducted on the Humaneval+ test dataset following He et al.~\cite{He2025RaSA}, which samples 50 completions per problem using a temperature of 0.2. We report Pass@1, Pass@5, and Pass@10 accuracy following standard protocol via BigCode Evaluation Harness~\cite{bigcode-evaluation-harness}. 

\textbf{Commonsense Reasoning} We fine-tuned LLaMA3.2–3B, LLaMA3.1–70B, Qwen-2.5-1.5B, and Qwen-2.5-7B (~\cite{yang2024qwen2}) across 8 commonsense tasks: BoolQ~\cite{clark2019boolq}, PIQA~\cite{bisk2020piqa}, SIQA~\cite{sap2019socialiqa}, HellaSwag~\cite{zellers2019hellaswag}, WinoGrande~\cite{sakaguchi2021winogrande}, ARC-Challenge, ARC-Easy~\cite{clark2018arc}, and OpenBookQA~\cite{mihaylov2018obqa}. We followed the training pipeline proposed by LLM-Adapters~\cite{hu2023llm}.
Training was performed on 2 H100 80G GPUs for 1.5-8B models, and on 8 A100 80G GPUs for the 70B model. LLaMA3.1–70B, Qwen-2.5-1.5B, and Qwen-2.5-7B were trained with rank 64, using the optimal configurations proposed by Biderman et al.~\cite{Biderman2024LoRALLFL}. LLaMA3.2–3B was trained with rank 32, following the settings of Ponkshe et al.~\cite{ponkshe2025initializationusingupdateapproximation} to ensure a fair comparison with results reported in the original paper. 

\textbf{Mathematical Reasoning} We fine-tuned LLaMA3.2–3B on MetaMathQA~\cite{yu2023metamath} train dataset using 4 H100 80G GPUs. Evaluation was done on MATH~\cite{hendrycks2021measuringmathematicalproblemsolving} dataset, following the evaluation procedure and settings from He et al.~\cite{He2025RaSA}. 

\textbf{General Language Understanding} We trained and evaluated RoBERTa-base~\cite{liu2019robertarobustlyoptimizedbert}, an encoder-only architecture model, on the GLUE~\cite{wang2019gluemultitaskbenchmarkanalysis} benchmark composed of eight sub-tasks. Following the protocol from prior works (~\cite{Kopiczko2024VeRA} ~\cite{gao2024parameterefficientfinetuningdiscretefourier}), we excluded MNLI and QQP—two time-intensive tasks—which also meant we did not apply the MNLI-based tricks for MRPC, RTE, and STS-B (as used in the original LoRA paper). Accordingly, we retrained LoRA on these tasks without this optimization and report updated results. All trainings were done on a single H100 80G GPU. 

\textbf{Personalized Image Generation} We fine-tuned SDXL~\cite{podell2023sdxlimprovinglatentdiffusion} following the official training setup from Huggingface diffusers repository, using the Naruto-Blip-Captions~\cite{cervenka2022naruto2} dataset on a single H100 80G GPU. The dataset was split 90\% for training and 10\% for evaluation. The quality was measured through CLIP similarity and DINOv2 similarity scores.

\textbf{Training Details}
We conducted experiments on five open-sourced LLMs—LLaMA3.1–8B, LLaMA3.1–70B, LLaMA3.2–3B, Qwen-2.5-1.5B, and Qwen-2.5-7B—covering diverse architecture and sclaes across code generation, commonsense reasoning, and mathematical reasoning tasks. Following common practice (~\cite{Jiang2024MoRA, Kopiczko2024VeRA}), we used pre-trained models rather than instruction-tuned models. All PEFT methods were applied to the linear modules in both the attention ( \(W_q, W_k, W_v, W_o\) )and the feed-forward networks (\( W_{up}, W_{down}, W_{gate}\)). We adopted alpaca-chat instruction template for training and evaluation. We compared GraLoRA to three representative PEFT methods: LoRA, MoRA ~\cite{Jiang2024MoRA} and RaSA ~\cite{He2025RaSA}.
We have also handled RoBERTa-base and SDXL, to show the robustness and scalability of our method across differnt models and tasks. Hyperparameters for GraLoRA followed those introduced in Kopiczko et al.~\cite{Kopiczko2024VeRA}, except for learning rate, which was reduced by a factor of 5–10, as VeRA uses a learning rate approximately 10 times larger than LoRA. Detailed training parameters can be found in Appendix~\ref{sec:train-details}. 

\subsection{Results on Code Generation}

\begin{table}[t]
\small
    \centering
    \caption{Pass@1, Pass@5, and Pass@10 results on LLaMA3.1–8B using LoRA, MoRA, RaSA, and GraLoRA across different ranks. Best results per group are in bold. * indicates Hybrid GraLoRA.}
    \resizebox{0.8\textwidth}{!}{
    \begin{tabular}{l c c c c c c}
        \toprule
        \textbf{Rank} & \textbf{Method} & \textbf{Training Time} & \textbf{Relative Time} & \textbf{Pass@1} & \textbf{Pass@5} & \textbf{Pass@10} \\
        \midrule
        \multirow{4}{*}{16}  & LoRA        & 6.2h    & 1.00\(\times\)  & 56.1\% & 65.3\% & 68.1\% \\
                             & MoRA        & 8.8h    & 1.42\(\times\)  & 53.6\% & 62.2\% & 64.5\% \\
                             & RaSA        & 6.7h    & 1.08\(\times\)  & 53.7\% & 64.4\% & 66.7\% \\
                             & GraLoRA*    & 6.7h    & 1.08\(\times\)  & \textbf{58.0\%} & \textbf{67.1}\% & \textbf{70.1\%} \\
        \midrule
        \multirow{4}{*}{32}  & LoRA        & 6.5h    & 1.00\(\times\)  & 58.4\% & \textbf{68.0}\% & 69.9\% \\
                             & MoRA        & 9.1h    & 1.40\(\times\)  & 58.3\% & 66.7\% & 69.0\% \\
                             & RaSA        & 6.8h    & 1.05\(\times\)  & 57.2\% & 67.9\% & \textbf{70.5}\% \\
                             & GraLoRA     & 6.9h    & 1.06\(\times\)  & \textbf{58.9\%} & 67.0\% & 69.0\% \\
        \midrule
        \multirow{4}{*}{64}  & LoRA        & 6.7h    & 1.00\(\times\)  & 58.1\% & 66.4\% & 68.5\% \\
                             & MoRA        & 9.7h    & 1.45\(\times\)  & 57.2\% & 66.4\% & 69.2\% \\
                             & RaSA        & 6.9h    & 1.03\(\times\)  & 56.6\% & 65.4\% & 67.9\% \\
                             & GraLoRA     & 7.2h    & 1.07\(\times\)  & \textbf{60.5\%} & \textbf{71.2}\% & \textbf{72.6\%} \\
        \midrule
        \multirow{4}{*}{128} & LoRA        & 7.0h    & 1.00\(\times\)  & 55.8\% & 64.8\% & 68.6\% \\
                             & MoRA        & 9.9h    & 1.41\(\times\)  & 52.8\% & 62.3\% & 65.3\% \\
                             & RaSA        & 7.6h    & 1.09\(\times\)  & 57.5\% & 65.5\% & 67.5\% \\
                             & GraLoRA     & 7.7h    & 1.10\(\times\)  & \textbf{64.3\%} & \textbf{71.7}\% & \textbf{73.7\%} \\
        \bottomrule
    \end{tabular}
    }
    \label{tab:code_generation}
\end{table}

As shown in Table~\ref{tab:code_generation}, GraLoRA outperformed LoRA, MoRA, and RaSA across all tested ranks for Pass@1 accuracy. At rank 64, GraLoRA achieved an absolute improvement of +2.4\% in Pass@1, +4.8\% in Pass@5, and +4.1\% in Pass@10 over LoRA. At rank 128, the gains were even more pronounced, with increases of +8.5\% in Pass@1, +6.9\% in Pass@5, and +5.1\% in Pass@10. Notably, while other methods struggled to fully utilize the increasing rank capacity—often reaching performance plateaus at lower ranks—GraLoRA maintained a consistent upward trajectory, effectively overcoming the limitations of LoRA.

Even in low-rank settings (e.g., rank 16), where expressive capacity is typically constrained, the hybrid variant of GraLoRA demonstrated superior performance. These improvements highlight GraLoRA’s enhanced capability to preserve diverse gradient signals and resist suppression from dominant outliers. The strong results on the HumanEval+ benchmark further underscore the benefits of fine-grained adaptation in tackling complex, high-precision code generation tasks.

\subsection{Results on Commonsense Reasoning}

\begin{table}[t]
\small
    \centering
    \caption{Commonsense reasoning accuracy across models and tasks. Bold indicates the best performance per column. HS means HellaSwag, and WG WinoGrande. \(^\dagger\) indicates values reported by Ponkshe et al.~\cite{ponkshe2025initializationusingupdateapproximation}}
    \resizebox{\textwidth}{!}{%
    \begin{tabular}{l l c c c c c c c c c}
        \toprule
        \textbf{Model} & \textbf{Method} & \textbf{BoolQ} & \textbf{PIQA} & \textbf{SIQA} & \textbf{HS} & \textbf{WG} & \textbf{ARC-c} & \textbf{ARC-e} & \textbf{OBQA} & \textbf{Avg.} \\
        \midrule
        \multirow{4}{*}{Qwen2.5-1.5B}
          & LoRA     & 66.5\% & 84.0\% & 74.9\% & 83.6\% & 73.7\% & 75.2\% & 88.1\% & 83.4\% & 78.7\% \\
          & MoRA     & 65.9\% & 82.2\% & 74.7\% & 82.6\% & 73.4\% & 72.6\% & 86.5\% & 82.8\% & 77.6\% \\
          & RaSA     & \textbf{67.5\%} & 83.7\% & 75.7\% & 85.3\% & 72.9\% & 76.4\% & 89.8\% & 83.8\% & 79.4\% \\
          & GraLoRA  & 67.2\% & \textbf{84.2\%} & \textbf{75.9\%} & \textbf{85.7\%} & \textbf{73.8\%} & \textbf{77.5\%} & \textbf{89.9\%} & \textbf{84.4\%} & \textbf{79.8\%} \\
        \midrule
        \multirow{4}{*}{Qwen2.5-7B}
          & LoRA$^{\dagger}$     & 72.3\% & 88.2\% & \textbf{79.2\%} & 92.9\% & \textbf{84.7\%} & 84.0\% & 93.6\% & 89.6\% & 85.6\% \\
          & MoRA     & 69.9\% & 85.3\% & 78.5\% & 83.7\% & 81.4\% & 77.5\% & 88.6\% & 85.0\% & 81.2\% \\
          & RaSA     & 72.0\% & 88.5\% & 78.9\% & \textbf{93.6\%} & 81.8\% & 86.1\% & 94.2\% & 90.2\% & 85.7\% \\
          & GraLoRA  & \textbf{73.4\%} & \textbf{89.7\%} & 79.0\% & 93.0\% & 84.0\% & \textbf{86.9\%} & \textbf{94.5\%} & \textbf{90.6\%} & \textbf{86.4\%} \\
        \midrule
        \multirow{4}{*}{LLaMA3.2–3B}
          & LoRA     & 70.0\% & 85.2\% & 79.1\% & 90.7\% & 82.2\% & 74.3\% & 86.9\% & 81.9\% & 81.3\% \\
          & MoRA     & 72.4\% & 86.1\% & 80.1\% & 92.3\% & 84.8\% & 76.8\% & 88.8\% & \textbf{84.8\%} & 83.3\% \\
          & RaSA     & 73.1\% & \textbf{87.5\%} & \textbf{81.1\%} & 93.7\% & 85.3\% & 78.9\% & 88.9\% & 83.6\% & 84.0\% \\
          & GraLoRA  & \textbf{74.1\%} & 86.5\% & 80.8\% & \textbf{93.8\%} & \textbf{87.5\%} & \textbf{79.9\%} & \textbf{89.5\%} & \textbf{84.8\%} & \textbf{84.6\%} \\
        \midrule
        \multirow{2}{*}{LLaMA3.1–70B}
          & LoRA     & 81.7\% & 93.4\% & 82.2\% & 97.5\% & 93.1\% & 90.2\% & 96.5\% & 95.6\% & 91.3\% \\
          & GraLoRA  & \textbf{83.1}\% & \textbf{94.7\%} & \textbf{83.6\%} & \textbf{97.9\%} & \textbf{93.8\%} & \textbf{92.3\%} & \textbf{97.8\%} & \textbf{96.2\%} & \textbf{92.4\%} \\
        \bottomrule
    \end{tabular}
    }
    \label{tab:commonsense_reasoning}
\end{table}

As shown in Table~\ref{tab:commonsense_reasoning}, GraLoRA outperformed other methods across a wide range of models and tasks. Notably, GraLoRA demonstrated superior performance across models of varying scales, achieving a 1.1\% improvement in average accuracy on both Qwen2.5-1.5B and LLaMA3.1-70B. It also yielded a 0.9\% gain on the widely used mid-sized model, Qwen2.5-7B. Moreover, GraLoRA achieved a 3.3\% improvement on LLaMA3.2-3B, surpassing a broad range of baselines as presented in Table~\ref{tab:commonsense_reasoning_additional}.

Furthermore, GraLoRA achieved the best results on 26 out of 32 tasks, consistently outperforming alternatives across benchmarks. These results support our analysis in Section~\ref{sec:gralora-gradients}, showing that GraLoRA’s localized updates enhance alignment with FFT and promote robust generalization in multi-aspect reasoning tasks.

\subsection{Results on Mathematical Reasoning}

\begin{table}[t]
\small
    \centering
    \vspace{-0.3cm}
    \caption{MATH dataset accuracy results on Qwen2.5-1.5B using LoRA and GraLoRA across different ranks. Best results per group are in bold.}
    \resizebox{0.6\textwidth}{!}{
    \begin{tabular}{l c c c c c c}
        \toprule
        \textbf{Rank} & \textbf{Method} & \textbf{Training Time} & \textbf{Relative Time} & \textbf{Accuracy} \\
        \midrule
        \multirow{2}{*}{64}  & LoRA       & 5.3h    & 1.00\(\times\)  & 23.6\% \\
                             & GraLoRA    & 6.2h    & 1.17\(\times\)  & 25.7\% \\
        \midrule
        \multirow{2}{*}{128} & LoRA       & 5.5h    & 1.00\(\times\)  & 24.7\% \\
                             & GraLoRA    & 6.6h    & 1.20\(\times\)  & 28.9\% \\
        \bottomrule
    \end{tabular}
    }
    \label{tab:math}
    \vspace{-0.3cm}
\end{table}

In mathematical reasoning task, regarded as one of the most challenging benchmarks, GraLoRA consistently outperformed LoRA across all configurations. Notably, in the high rank setting of \( r = 128 \), GraLoRA achieved a 4.2\% improvement in accuracy (Table~\ref{tab:math}), mirroring the performance trends observed in the code generation experiments. These results further highlight the robustness of GraLoRA, demonstrating its capability to fully exploit the advantages enabled by increased rank capacity, thereby overcoming the inherent expressiveness constraints of previous PEFT methods.

\subsection{Results on General Language Understanding}

GraLoRA demonstrates strong performance even in the low-rank regime, outperforming all baselines in terms of average score. The Hybrid GraLoRA variant achieves the most robust results, attaining the best performance on four out of six tasks, while both the original and hybrid versions consistently surpass all other baselines, as shown in Table~\ref{tab:glue}. Compared with LoRA, the best GraLoRA configuration yields a 1.8\% improvement in average accuracy, with gains observed across all sub-tasks. These findings indicate that GraLoRA maintains high effectiveness even under constrained parameter budgets and generalizes well to non-LLM architectures.

\subsection{Results on Personalized Image Generation}

In the image generation task, GraLoRA consistently outperformed LoRA in both CLIP and DINOv2 similarity metrics, achieving 0.5\% and 2.1\% improvements, respectively (Table~\ref{tab:vision}). These results further demonstrate the generality and effectiveness of GraLoRA beyond language models, extending its applicability to vision–language and generative architectures such as diffusion models.

\subsection{Ablation Study}
\label{sec:ablation}

\begin{figure}[t]
    \centering
    \includegraphics[width=1.0\linewidth]{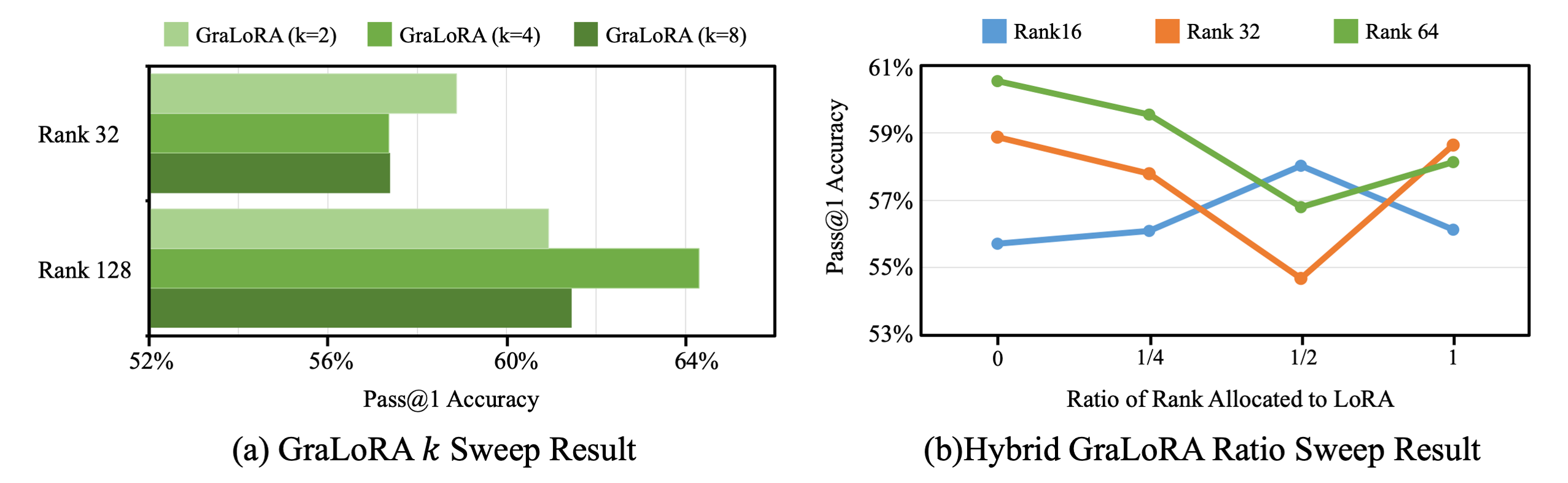}
    \caption{(a) GraLoRA \( k \) sweep results and (b) Hybrid GraLoRA Ratio sweep results for LLaMA3.1–8B on code generation task. Ratio 0 implies default GraLoRA and ratio 1 vanilla LoRA in (b).}
    \label{fig:param_sweep}
    \vspace{-0.3cm}
\end{figure}

\textbf{GraLoRA \( k \) Sweep} \quad We evaluated the impact of varying \( k \) on code generation accuracy. As shown in Figure~\ref{fig:param_sweep} (a), \( k=2 \) yielded the best performance at rank 32, while \( k=4 \) was optimal at rank 128. These results are consistent with the theoretical prediction that a smaller \( k \) is preferable for lower ranks, as reduced sub-block rank can be particularly detrimental when the overall rank is limited.

\textbf{Hybrid GraLoRA Ratio Sweep} \quad We assessed performance across different LoRA-to-GraLoRA rank allocation ratios for the Hybrid GraLoRA configuration (Figure ~\ref{fig:param_sweep} (b)). At rank 16, partially allocating the rank to LoRA led optimal accuracy. However, for larger ranks, allocating rank to LoRA resulted in degraded performance. This suggests that Hybrid GraLoRA is advantageous in low-rank regimes, where the sub-block rank of GraLoRA alone may be insufficient. In contrast, under higher-rank settings where GraLoRA’s sub-blocks are expressive enough, introducing LoRA components may lead to gradient entanglement, thereby hindering effective learning.

\section{Conclusion}
\label{sec:conclusion}

In this work, we introduced \textbf{GraLoRA}, a novel PEFT method that extends LoRA with granular, block-wise decomposition. Motivated by a rigorous analysis of LoRA's gradient behavior, we identified that input outliers can dominate the low-rank update, suppressing meaningful contributions from other input channels and misaligning with the localized gradient propagation observed in FFT.

GraLoRA addresses this limitation by dividing the adaptation space into $k^2$ independently trained low-rank adapters, enabling spatially localized and context-aware updates. Our theoretical analysis shows that this design increases expressivity by a factor of $k$, without additional parameters or computational cost. Moreover, under outlier activations, GraLoRA effectively mitigates the global gradient distortion seen in vanilla LoRA and better preserves inter-channel balance. Empirically, GraLoRA consistently outperforms standard LoRA and strong baselines such as RaSA across diverse tasks and model scales. On the code generation benchmark HumanEval+, it achieves up to +8.5\% absolute gain in Pass\@1. GraLoRA also delivers significant improvements across other 4 additional tasks, highlighting its robustness and scalability across heterogeneous architectures and model sizes. 

\paragraph{Future Work.} While GraLoRA improves gradient locality and expressive power, its current design assumes uniform partitioning. Future extensions may explore adaptive or learned partitioning schemes, sparsity-aware block activation, or task-driven dynamic rank allocation. Additionally, applying GraLoRA to vision transformers, multimodal architectures, or continual learning setups may further highlight its potential for robust and efficient model adaptation.

Overall, GraLoRA represents a principled and practical step forward in the design of PEFT methods, bridging the gap between global low-rank reparameterization and local, fine-grained adaptation.

\newpage
\begin{ack}
This work was partly supported by the Technology development Program of MSS [RS-2023-00258531], the Starting growth Technological R\&D Program of MSS [S3358777], and the Institute of Information \& communications Technology Planning \& Evaluation (IITP) grant funded by the Korea government (MSIT) (No. RS-2025-02304183, Development of Optimization Code Conversion Technology for Heterogeneous Al Semiconductor-Based Large-Scale Models).
\end{ack}

\bibliographystyle{plain}
\bibliography{references}

@article{yang2024qwen2,
  title={Qwen2. 5 technical report},
  author={Yang, An and Yang, Baosong and Zhang, Beichen and Hui, Binyuan and Zheng, Bo and Yu, Bowen and Li, Chengyuan and Liu, Dayiheng and Huang, Fei and Wei, Haoran and others},
  journal={arXiv preprint arXiv:2412.15115},
  year={2024}
}

@misc{grattafiori2024llama3herdmodels,
      title={The Llama 3 Herd of Models}, 
      author={Aaron Grattafiori and others},
      year={2024},
      eprint={2407.21783},
      archivePrefix={arXiv},
      primaryClass={cs.AI},
      url={https://arxiv.org/abs/2407.21783}, 
}

@misc{houlsby2019parameterefficienttransferlearningnlp,
      title={Parameter-Efficient Transfer Learning for NLP}, 
      author={Neil Houlsby and Andrei Giurgiu and Stanislaw Jastrzebski and Bruna Morrone and Quentin de Laroussilhe and Andrea Gesmundo and Mona Attariyan and Sylvain Gelly},
      year={2019},
      eprint={1902.00751},
      archivePrefix={arXiv},
      primaryClass={cs.LG},
      url={https://arxiv.org/abs/1902.00751}, 
}

@article{Hu2021LoRA,
  author  = {Edward Hu and Yelong Shen and Phillip Wallis and
             Zeyuan Allen-Zhu and Yuanzhi Li and Shean Wang and
             Lu Wang and Weizhu Chen},
  title   = {{LoRA}: Low-Rank Adaptation of Large Language Models},
  journal = {arXiv preprint arXiv:2106.09685},
  year    = {2021}
}

@misc{xu2023peft_methods,
      title={Parameter-Efficient Fine-Tuning Methods for Pretrained Language Models: A Critical Review and Assessment}, 
      author={Lingling Xu and Haoran Xie and Si-Zhao Joe Qin and Xiaohui Tao and Fu Lee Wang},
      year={2023},
      eprint={2312.12148},
      archivePrefix={arXiv},
      primaryClass={cs.CL},
      url={https://arxiv.org/abs/2312.12148}, 
}

@article{Jiang2024MoRA,
  author  = {Ting Jiang and Shaohan Huang and Shengyue Luo and Zihan Zhang and
             Haizhen Huang and Furu Wei and Weiwei Deng and Feng Sun and
             Qi Zhang and Deqing Wang and Fuzhen Zhuang},
  title   = {{MoRA}: High‑Rank Updating for Parameter‑Efficient Fine‑Tuning},
  journal = {arXiv preprint arXiv:2405.12130},
  year    = {2024}
}

@inproceedings{He2025RaSA,
  author    = {Zhiwei He and Zhaopeng Tu and Xing Wang and Xingyu Chen and
               Zhijie Wang and Jiahao Xu and Tian Liang and Wenxiang Jiao and
               Zhuosheng Zhang and Rui Wang},
  title     = {{RaSA}: Rank‑Sharing Low‑Rank Adaptation},
  booktitle = {Proceedings of the 2025 International Conference on Learning
               Representations (ICLR)},
  year      = {2025}
}

@inproceedings{Kopiczko2024VeRA,
  author    = {Dawid J. Kopiczko and Tijmen Blankevoort and Yuki M. Asano},
  title     = {{VeRA}: Vector‑Based Random Matrix Adaptation},
  booktitle = {Proceedings of the 2024 International Conference on Learning
               Representations (ICLR)},
  year      = {2024}
}

@inproceedings{Huang2025HiRA,
  author    = {Qiushi Huang and Tom Ko and Zhan Zhuang and Lilian Tang and
               Yu Zhang},
  title     = {{HiRA}: Parameter‑Efficient Hadamard High‑Rank Adaptation for
               Large Language Models},
  booktitle = {Proceedings of the 2025 International Conference on Learning
               Representations (ICLR)},
  year      = {2025}
}

@article{Biderman2024LoRALLFL,
  author  = {Dan Biderman and Jacob Portes and Jose Javier Gonzalez Ortiz and
             Mansheej Paul and Philip Greengard and Connor Jennings and
             Daniel King and Sam Havens and Vitaliy Chiley and Jonathan Frankle and
             Cody Blakeney and John~P. Cunningham},
  title   = {{LoRA} Learns Less and Forgets Less},
  journal = {Transactions on Machine Learning Research},
  year    = {2024}
}

@misc{xiao2024smoothquantaccurateefficientposttraining,
      title={SmoothQuant: Accurate and Efficient Post-Training Quantization for Large Language Models}, 
      author={Guangxuan Xiao and Ji Lin and Mickael Seznec and Hao Wu and Julien Demouth and Song Han},
      year={2024},
      eprint={2211.10438},
      archivePrefix={arXiv},
      primaryClass={cs.CL},
      url={https://arxiv.org/abs/2211.10438}, 
}

@article{paischer2024one,
  title={One initialization to rule them all: Fine-tuning via explained variance adaptation},
  author={Paischer, Fabian and Hauzenberger, Lukas and Schmied, Thomas and Alkin, Benedikt and Deisenroth, Marc Peter and Hochreiter, Sepp},
  journal={arXiv preprint arXiv:2410.07170},
  year={2024}
}

@article{meng2024pissa,
  title={Pissa: Principal singular values and singular vectors adaptation of large language models},
  author={Meng, Fanxu and Wang, Zhaohui and Zhang, Muhan},
  journal={Advances in Neural Information Processing Systems},
  volume={37},
  pages={121038--121072},
  year={2024}
}

@article{buyukakyuz2024olora,
  title={Olora: Orthonormal low-rank adaptation of large language models},
  author={B{\"u}y{\"u}kaky{\"u}z, Kerim},
  journal={arXiv preprint arXiv:2406.01775},
  year={2024}
}

@article{wang2024lora,
  title={Lora-ga: Low-rank adaptation with gradient approximation},
  author={Wang, Shaowen and Yu, Linxi and Li, Jian},
  journal={Advances in Neural Information Processing Systems},
  volume={37},
  pages={54905--54931},
  year={2024}
}

@misc{bigcode-evaluation-harness,
  author       = {Ben Allal, Loubna and
                  Muennighoff, Niklas and
                  Kumar Umapathi, Logesh and
                  Lipkin, Ben and
                  von Werra, Leandro},
  title = {A framework for the evaluation of code generation models},
  publisher = {GitHub},
  journal = {GitHub repository},
  howpublished = {\url{https://github.com/bigcode-project/bigcode-evaluation-harness}},
  year = 2022,
}

@inproceedings{hu2023llm,
  title={LLM-Adapters: An Adapter Family for Parameter-Efficient Fine-Tuning of Large Language Models},
  author={Hu, Zhiqiang and Wang, Lei and Lan, Yihuai and Xu, Wanyu and Lim, Ee-Peng and Bing, Lidong and Xu, Xing and Poria, Soujanya and Lee, Roy},
  booktitle={Proceedings of the 2023 Conference on Empirical Methods in Natural Language Processing},
  pages={5254--5276},
  year={2023}
}

@inproceedings{lee2024owq,
  title={Owq: Outlier-aware weight quantization for efficient fine-tuning and inference of large language models},
  author={Lee, Changhun and Jin, Jungyu and Kim, Taesu and Kim, Hyungjun and Park, Eunhyeok},
  booktitle={Proceedings of the AAAI Conference on Artificial Intelligence},
  volume={38},
  number={12},
  pages={13355--13364},
  year={2024}
}

@InProceedings{wei2024magicoder,
  title = 	 {Magicoder: Empowering Code Generation with {OSS}-Instruct},
  author =       {Wei, Yuxiang and Wang, Zhe and Liu, Jiawei and Ding, Yifeng and Zhang, Lingming},
  booktitle = 	 {Proceedings of the 41st International Conference on Machine Learning},
  pages = 	 {52632--52657},
  year = 	 {2024},
  editor = 	 {Salakhutdinov, Ruslan and Kolter, Zico and Heller, Katherine and Weller, Adrian and Oliver, Nuria and Scarlett, Jonathan and Berkenkamp, Felix},
  volume = 	 {235},
  series = 	 {Proceedings of Machine Learning Research},
  month = 	 {21--27 Jul},
  publisher =    {PMLR},
  url = 	 {https://proceedings.mlr.press/v235/wei24h.html}
}

@inproceedings{luo2024wizardcoder,
    title={WizardCoder: Empowering Code Large Language Models with Evol-Instruct},
    author={Ziyang Luo and Can Xu and Pu Zhao and Qingfeng Sun and Xiubo Geng and Wenxiang Hu and Chongyang Tao and Jing Ma and Qingwei Lin and Daxin Jiang},
    booktitle={The Twelfth International Conference on Learning Representations},
    year={2024},
    url={https://openreview.net/forum?id=UnUwSIgK5W}
}

@inproceedings{clark2019boolq,
    title = "{B}ool{Q}: Exploring the Surprising Difficulty of Natural Yes/No Questions",
    author = "Clark, Christopher  and
      Lee, Kenton  and
      Chang, Ming-Wei  and
      Kwiatkowski, Tom  and
      Collins, Michael  and
      Toutanova, Kristina",
    editor = "Burstein, Jill  and
      Doran, Christy  and
      Solorio, Thamar",
    booktitle = "Proceedings of the 2019 Conference of the North {A}merican Chapter of the Association for Computational Linguistics: Human Language Technologies, Volume 1 (Long and Short Papers)",
    month = jun,
    year = "2019",
    address = "Minneapolis, Minnesota",
    publisher = "Association for Computational Linguistics",
    url = "https://aclanthology.org/N19-1300/",
    doi = "10.18653/v1/N19-1300",
    pages = "2924--2936"
}

@inproceedings{bisk2020piqa,
  title={Piqa: Reasoning about physical commonsense in natural language},
  author={Bisk, Yonatan and Zellers, Rowan and Gao, Jianfeng and Choi, Yejin and others},
  booktitle={Proceedings of the AAAI conference on artificial intelligence},
  volume={34},
  number={05},
  pages={7432--7439},
  year={2020}
}

@article{sap2019socialiqa,
  title={Socialiqa: Commonsense reasoning about social interactions},
  author={Sap, Maarten and Rashkin, Hannah and Chen, Derek and LeBras, Ronan and Choi, Yejin},
  journal={arXiv preprint arXiv:1904.09728},
  year={2019}
}

@article{zellers2019hellaswag,
  title={Hellaswag: Can a machine really finish your sentence?},
  author={Zellers, Rowan and Holtzman, Ari and Bisk, Yonatan and Farhadi, Ali and Choi, Yejin},
  journal={arXiv preprint arXiv:1905.07830},
  year={2019}
}

@article{sakaguchi2021winogrande,
  title={Winogrande: An adversarial winograd schema challenge at scale},
  author={Sakaguchi, Keisuke and Bras, Ronan Le and Bhagavatula, Chandra and Choi, Yejin},
  journal={Communications of the ACM},
  volume={64},
  number={9},
  pages={99--106},
  year={2021},
  publisher={ACM New York, NY, USA}
}

@article{clark2018arc,
  title={Think you have solved question answering? try arc, the ai2 reasoning challenge},
  author={Clark, Peter and Cowhey, Isaac and Etzioni, Oren and Khot, Tushar and Sabharwal, Ashish and Schoenick, Carissa and Tafjord, Oyvind},
  journal={arXiv preprint arXiv:1803.05457},
  year={2018}
}

@misc{mihaylov2018obqa,
      title={Can a Suit of Armor Conduct Electricity? A New Dataset for Open Book Question Answering}, 
      author={Todor Mihaylov and Peter Clark and Tushar Khot and Ashish Sabharwal},
      year={2018},
      eprint={1809.02789},
      archivePrefix={arXiv},
      primaryClass={cs.CL},
      url={https://arxiv.org/abs/1809.02789}, 
}

@article{yu2023metamath,
  title={MetaMath: Bootstrap Your Own Mathematical Questions for Large Language Models},
  author={Yu, Longhui and Jiang, Weisen and Shi, Han and Yu, Jincheng and Liu, Zhengying and Zhang, Yu and Kwok, James T and Li, Zhenguo and Weller, Adrian and Liu, Weiyang},
  journal={arXiv preprint arXiv:2309.12284},
  year={2023}
}

@misc{hendrycks2021measuringmathematicalproblemsolving,
      title={Measuring Mathematical Problem Solving With the MATH Dataset}, 
      author={Dan Hendrycks and Collin Burns and Saurav Kadavath and Akul Arora and Steven Basart and Eric Tang and Dawn Song and Jacob Steinhardt},
      year={2021},
      eprint={2103.03874},
      archivePrefix={arXiv},
      primaryClass={cs.LG},
      url={https://arxiv.org/abs/2103.03874}, 
}

@misc{liu2019robertarobustlyoptimizedbert,
      title={RoBERTa: A Robustly Optimized BERT Pretraining Approach}, 
      author={Yinhan Liu and Myle Ott and Naman Goyal and Jingfei Du and Mandar Joshi and Danqi Chen and Omer Levy and Mike Lewis and Luke Zettlemoyer and Veselin Stoyanov},
      year={2019},
      eprint={1907.11692},
      archivePrefix={arXiv},
      primaryClass={cs.CL},
      url={https://arxiv.org/abs/1907.11692}, 
}

@misc{wang2019gluemultitaskbenchmarkanalysis,
      title={GLUE: A Multi-Task Benchmark and Analysis Platform for Natural Language Understanding}, 
      author={Alex Wang and Amanpreet Singh and Julian Michael and Felix Hill and Omer Levy and Samuel R. Bowman},
      year={2019},
      eprint={1804.07461},
      archivePrefix={arXiv},
      primaryClass={cs.CL},
      url={https://arxiv.org/abs/1804.07461}, 
}

@misc{gao2024parameterefficientfinetuningdiscretefourier,
      title={Parameter-Efficient Fine-Tuning with Discrete Fourier Transform}, 
      author={Ziqi Gao and Qichao Wang and Aochuan Chen and Zijing Liu and Bingzhe Wu and Liang Chen and Jia Li},
      year={2024},
      eprint={2405.03003},
      archivePrefix={arXiv},
      primaryClass={cs.LG},
      url={https://arxiv.org/abs/2405.03003}, 
}

@misc{podell2023sdxlimprovinglatentdiffusion,
      title={SDXL: Improving Latent Diffusion Models for High-Resolution Image Synthesis}, 
      author={Dustin Podell and Zion English and Kyle Lacey and Andreas Blattmann and Tim Dockhorn and Jonas Müller and Joe Penna and Robin Rombach},
      year={2023},
      eprint={2307.01952},
      archivePrefix={arXiv},
      primaryClass={cs.CV},
      url={https://arxiv.org/abs/2307.01952}, 
}

@misc{cervenka2022naruto2,
      author = {Cervenka, Eole},
      title = {Naruto BLIP captions},
      year={2022},
      howpublished= {\url{https://huggingface.co/datasets/lambdalabs/naruto-blip-captions/}}
}

@misc{ponkshe2025initializationusingupdateapproximation,
      title={Initialization using Update Approximation is a Silver Bullet for Extremely Efficient Low-Rank Fine-Tuning}, 
      author={Kaustubh Ponkshe and Raghav Singhal and Eduard Gorbunov and Alexey Tumanov and Samuel Horvath and Praneeth Vepakomma},
      year={2025},
      eprint={2411.19557},
      archivePrefix={arXiv},
      primaryClass={cs.CL},
      url={https://arxiv.org/abs/2411.19557}, 
}

\newpage
\appendix
\section{Rank Analysis in Real-World Scenarios}
\label{sec:rank-real}

\begin{table}[H]
\centering
    \caption{Average rank size in each projection layer across LoRA and GraLoRA variants. Rank \(r\) was set to 128 in all methods.}
    \begin{tabular}{|c|c|c|c|c|c|c|c|}
        \hline
         & q\_proj & k\_proj & v\_proj & o\_proj & up\_proj & down\_proj & gate\_proj \\ \hline
        LoRA & 128 & 128 & 128 & 128 & 128 & 128 & 128 \\ \hline
        GraLoRA (k=2) & 256 & 256 & 256 & 256 & 256 & 256 & 256 \\ \hline
        GraLoRA (k=4) & 512 & 512 & 512 & 512 & 512 & 512 & 512 \\ \hline
        GraLoRA (k=8) & 1024 & 1016 & 1022 & 1024 & 1024 & 1024 & 1024 \\ \hline
        \end{tabular}
    \label{tab:rank_size}
\end{table}

As shown in Table~\ref{tab:rank_size}, GraLoRA denoted linearly increasing ranks as the \( k \) increased. The observation aligns with our theoretical analysis that increasing GraLoRA \( k \) leads to higher expression power by increasing the latent space from \( r \) to \( kr \).

\section{Gradient Distribution of LoRA and GraLoRA}
\label{sec:gradient_distribution}

\begin{figure}[H]
    \centering
    \includegraphics[width=1.0\linewidth]{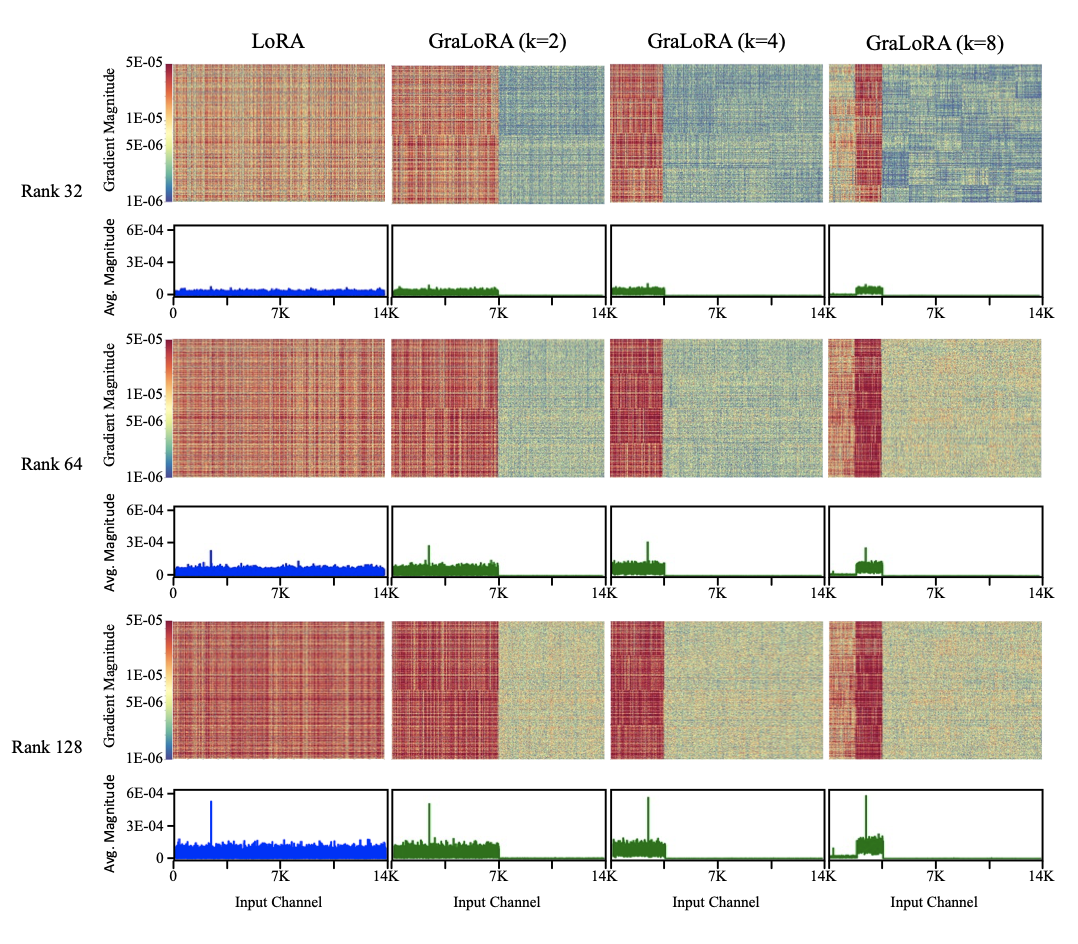}
    \caption{Comparison of gradient distributions under outlier activation for rank 32, 64, and 128 in LLaMA3.1-8B Layer 1 down-projection matrix.}
    \label{fig:gralora_grad_full}
\end{figure}

Figure~\ref{fig:gralora_grad_full} displays gradient distributions of LoRA and GraLoRA for varying ranks. In GraLoRA, only the blocks interacting with the outlier exhibit elevated gradients, structurally solving the gradient entanglement discovered in vanilla LoRA. This enables to mitigate global distortion and align with FFT behavior in all ranks.

\section{Precise Analysis on Computation Overhead}
\label{sec:compute}

\begin{figure}[H]
    \centering
    \includegraphics[width=0.8\linewidth]{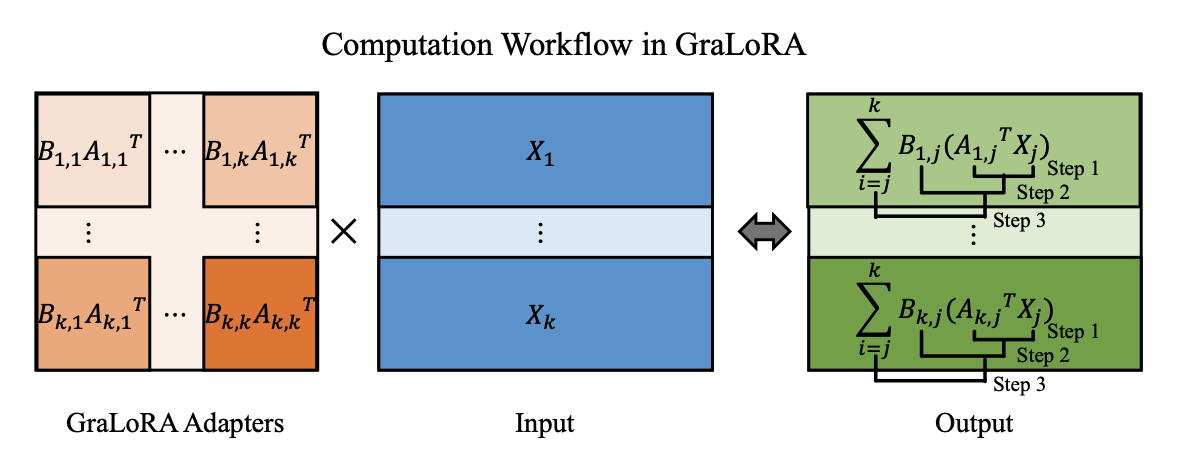}
    \caption{Computation workflow in GraLoRA is composed of 3 steps: two sub-block matrix multiplications and a following matrix addition.}
    \label{fig:computation_workflow}
\end{figure}

In the previous \textit{``Computation Overhead Analysis''} section~\ref{sec:tradeoff} we compared the computation of LoRA and GraLoRA with the big \(O\) notation on the two major matrix multiplication steps. In this section we further examine the exact computation requirement and compare their efficiency. 

\textbf{LoRA FLOPs} LoRA performs the projection in two sequential steps to leverage its low-rank structure. In the first step, the computation of \( A^\top X \in \mathbb{R}^{r \times T} \) requires \( (2N - 1)rT \) FLOPs. In the second step, the reconstruction \( B(A^\top X) \in \mathbb{R}^{M \times T} \) incurs \( (2r - 1)MT \) FLOPs. Therefore, the total FLOPs for LoRA is:
\begin{align*}
\text{LoRA}_{\text{FLOPs}} &= (2N - 1)rT + (2r - 1)MT \\
&= 2r(M + N)T - (r + M)T.
\end{align*}

\textbf{GraLoRA FLOPs} \quad In practice, GraLoRA computations can be divided into three stages, involving \( k^2 \) adapter blocks: two matrix multiplications followed by a matrix addition as shown in Figure~\ref{fig:computation_workflow}.

In the first stage (projection), each adapter block computes \( A_{i,j}^\top X_j \in \mathbb{R}^{\frac{r}{k} \times T} \), which requires \( \left(2\frac{n}{k} - 1\right)\frac{r}{k}T \) FLOPs. Since there are \( k^2 \) such blocks, the total FLOPs for this step is \( (2n - k)rT \).

In the second stage (reconstruction), each adapter block performs \( B_{i,j}(A_{i,j}^\top X_j) \in \mathbb{R}^{\frac{m}{k} \times T} \), which costs \( \left(2\frac{r}{k} - 1\right)\frac{m}{k}T \) FLOPs. With \( k^2 \) blocks, the total becomes \( (2r - k)mT \).

The final stage involves aggregating the outputs across \( k \) projections for each row:
\[
\sum_{j=1}^k B_{i,j}(A_{i,j}^\top X_j) \in \mathbb{R}^{\frac{m}{k} \times T},
\]
which requires \( \left(\frac{m}{k} \times T\right)(k - 1) = \frac{mT(k - 1)}{k} \) FLOPs per row. Across \( k \) rows, the total cost becomes \( (k - 1)mT \).

Combining all three stages, the total FLOPs for GraLoRA is:
\begin{align*}
\text{GraLoRA}_{\text{FLOPs}} &= (2n - k)rT + (2r - k)mT + (k - 1)mT \\
&= 2r(m + n)T - k(r + m)T + (k - 1)mT \\
&= 2r(m + n)T - krT - mT.
\end{align*}

This can also be expressed as:
\[
\text{GraLoRA}_{\text{FLOPs}} = \text{LoRA}_{\text{FLOPs}} - (k - 1)rT,
\]
demonstrating that GraLoRA introduces reduced computation compared to LoRA.

\section{Additional Experiment Results}
\label{sec:additional_baselines}

\subsection{Results on Commonsense Reasoning with extensive baseline comparison}

\begin{table}[H]
\small
    \centering
    \caption{Commonsense reasoning accuracy across models and tasks. All values are percentages; bold indicates the best performance per row. HS means HellaSwag, and WG WinoGrande. \(^\dagger\) indicates values reported by Ponkshe et al.~\cite{ponkshe2025initializationusingupdateapproximation}}
    \resizebox{\textwidth}{!}{
    \begin{tabular}{l l l | c c c c c c c c c}
        \toprule
        \textbf{Method} & \textbf{Rank} & \textbf{\#Parmas} & \textbf{BoolQ} & \textbf{PIQA} & \textbf{SIQA} & \textbf{HS} & \textbf{WG} & \textbf{ARC-c} & \textbf{ARC-e} & \textbf{OBQA} & \textbf{Avg.} \\
        \midrule
          Full-FT$^\dagger$  & - & 3.21B  & 70.4\% & 85.6\% & 80.5\% & 91.9\% & 85.0\% & 75.3\% & 88.5\% & 81.9\% & 82.4\% \\
        \midrule
          LoRA-XS$^\dagger$	 & 96 & 1.81M  & 67.3\%	& 83.4\% & 78.7\% &	89.0\% & 82.1\% & 72.6\% & 85.2\% &	78.9\% & 79.6\% \\
          LoRA-SB$^\dagger$	 & 96 & 1.81M  & 70.3\%	& 84.8\% & 80.2\% &	91.6\% & 84.6\% & 74.7\% & 87.9\% &	81.2\% & 81.9\% \\
        \midrule
          LoRA$^\dagger$     & 32 & 48.63M & 70.0\% & 85.2\% & 79.1\% & 90.7\% & 82.2\% & 74.3\% & 86.9\% & 81.9\% & 81.3\% \\
          MELoRA	 & 32 & 48.63M & 71.3\% & 85.0\% & 78.6\% &	93.0\% & 79.7\% & 73.7\% & 85.5\% &	79.0\% & 80.7\% \\
          rsLoRA$^\dagger$	 & 32 & 48.63M & 69.8\%	& 85.1\% & 78.9\% &	90.5\% & 82.0\% & 74.2\% & 86.7\% &	81.7\% & 81.1\% \\
          PiSSA$^\dagger$	 & 32 & 48.63M & 70.1\%	& 85.4\% & 79.4\% &	90.9\% & 82.7\% & 74.6\% & 87.2\% &	81.8\% & 81.5\% \\
          DoRA$^\dagger$     & 32 & 49.40M & 70.4\% & 85.6\% & 79.7\% & 90.8\% & 82.9\% & 74.9\% & 87.6\% & 82.0\% & 81.7\% \\
          BOFT	 & 32 & 48.48M & 72.3\% & 84.6\% & 79.1\% &	91.3\% & 84.5\% & 73.7\% & 87.8\% &	80.6\% & 81.7\% \\
          LoRA-Pro$^\dagger$ & 32 & 48.63M & 71.3\% & 85.8\% & 79.4\% & 90.9\% & 83.4\% & 75.3\% & 87.2\% & 81.7\% & 81.9\% \\
          MoRA     & 32 & 48.63M & 72.4\% & 86.1\% & 80.1\% & 92.3\% & 84.8\% & 76.8\% & 88.8\% & \textbf{84.8\%} & 83.3\% \\
          RaSA     & 32 & 48.63M & 73.1\% & \textbf{87.5\%} & \textbf{81.1\%} & 93.7\% & 85.3\% & 78.9\% & 88.9\% & 83.6\% & 84.0\% \\
          GraLoRA  & 32 & 48.63M & \textbf{74.1\%} & 86.5\% & 80.8\% & \textbf{93.8\%} & \textbf{87.5\%} & \textbf{79.9\%} & \textbf{89.5\%} & \textbf{84.8\%} & \textbf{84.6\%} \\
        \bottomrule
    \end{tabular}
    }
    \label{tab:commonsense_reasoning_additional}
\end{table}

\subsection{Results on General Language Understanding (GLUE)}

\begin{table}[H]
\small
    \centering
    \caption{GLUE dataset accuracy results on RoBERTa-base across different tasks. Best results per task are in bold.}
    \resizebox{\textwidth}{!}{
    \begin{tabular}{l l | c c c c c c c}
        \toprule
        \textbf{Method} & \textbf{\#Params} & \textbf{SST-2 (\%)} & \textbf{MRPC (\%)} & \textbf{CoLA (\%)} & \textbf{QNLI (\%)} & \textbf{RTE (\%)} & \textbf{STS-B (\%)} & \textbf{Avg (\%)} \\
        \midrule
        Full-FT        & 125M   & 94.8 & 90.2 & 63.6 & 92.8 & 78.7 & \textbf{91.2} & 85.2 \\
        LoRA           & 0.3M   & 95.1 & 86.5 & 63.4 & 93.3 & 76.2 & 90.6 & 84.2 \\
        VeRA           & 0.043M & 94.6 & 89.5 & 65.6 & 91.8 & 78.7 & 90.7 & 85.2 \\
        FourierFT      & 0.024M & 94.2 & 90.0 & 63.8 & 92.2 & 79.1 & 90.8 & 85.0 \\
        GraLoRA        & 0.3M   & \textbf{95.2} & 89.7 & \textbf{65.3} & 93.0 & \textbf{80.9} & 91.1 & 85.8 \\
        Hybrid GraLoRA & 0.3M   & \textbf{95.2} & \textbf{90.2} & 64.1 & \textbf{93.4} & 79.8 & \textbf{91.2} & 85.6 \\
        Best GraLoRA   & 0.3M   & \textbf{95.2} & \textbf{90.2} & \textbf{65.3} & \textbf{93.4} & \textbf{80.9} & \textbf{91.2} & \textbf{86.0} \\
        \bottomrule
    \end{tabular}
    }
    \label{tab:glue}
\end{table}

\subsection{Results on Personalized Image Generation}

\begin{table}[H]
\small
    \centering
    \caption{SDXL fine-tuning results personalized image generation.}
    \resizebox{0.6\textwidth}{!}{
    \begin{tabular}{l | c c}
        \toprule
        \textbf{Method} & \textbf{CLIP Similarity} & \textbf{DINOv2 Similarity} \\
        \midrule
        LoRA            & 91.4\% & 79.2\% \\
        GraLoRA         & \textbf{91.9\%} & \textbf{81.3\%} \\ 
        \bottomrule
    \end{tabular}
    }
    \label{tab:vision}
\end{table}

\section{Experiment Details}
\label{sec:train-details}
\textbf{Baseline Methods}. We compared GraLoRA with three main baseline methods. Key idea for each method is as follows:

\begin{itemize}
    \item \textbf{LoRA} freezes pretrained model weights and injects trainable low-rank matrices into selected layers, allowing efficient fine-tuning with significantly fewer parameters, approximating weight updates as a product of two small matrices.
    \item \textbf{MoRA} employs a single square matrix instead of low-rank matrices to achieve high-rank updating while maintaining the same number of trainable parameters.
    \item \textbf{RaSA} enhances LoRA by sharing partial low-rank components across layers while keeping layer-specific updates.
\end{itemize}

\begin{table}[ht]
\centering
\caption{Hyperparameters for Code Generation, Commonsense Reasoning, Mathematical Reasoning, and Personalized Image Generation tasks.}
\resizebox{\textwidth}{!}{
\begin{tabular}{ccccccccccc}
\toprule
\textbf{Task} & \textbf{Model} & \textbf{Method} & \textbf{Rank} & \textbf{LR} & \textbf{Batch size} & \textbf{Epochs} & \textbf{Optimizer}\\
\midrule

\multirow{4}{*}{\makecell{Code \\ Generation}} & \multirow{4}{*}{LLaMA3.1–8B} & LoRA & \multirow{4}{*}{\{16, 32, 64, 128\}} & \multirow{4}{*}{2e-4} & \multirow{4}{*}{192} & \multirow{4}{*}{2} & \multirow{4}{*}{LionW}\\
& & MoRA        & & & & & \\
& & RaSA        & & & & & \\
& & GraLoRA     & & & & & \\
\midrule

\multirow{15}{*}{\makecell{Commonsense \\ Reasoning}}
& \multirow{4}{*}{Qwen-2.5-1.5B} & LoRA & \multirow{4}{*}{64} & \multirow{4}{*}{2e-4} & \multirow{4}{*}{192} & \multirow{4}{*}{2} & \multirow{4}{*}{LionW} \\
& & MoRA        & & & & & \\
& & RaSA        & & & & & \\
& & GraLoRA     & & & & & \\
\cmidrule(l){2-8}
& \multirow{4}{*}{Qwen-2.5-7B} & LoRA & \multirow{4}{*}{64} & \multirow{4}{*}{4e-4} & \multirow{4}{*}{192} & \multirow{4}{*}{2} & \multirow{4}{*}{LionW} \\
& & MoRA        & & & & & \\
& & RaSA        & & & & & \\
& & GraLoRA     & & & & & \\
\cmidrule(l){2-8}
& \multirow{5}{*}{LLaMA3.2–3B} & BOFT & \multirow{5}{*}{32} & {4e-4} & \multirow{5}{*}{192} & \multirow{5}{*}{2} & \multirow{5}{*}{AdamW} \\
& & MeLORA      & & 4e-4 & & & \\
& & MoRA        & & 2e-4 & & & \\
& & RaSA        & & 4e-4 & & & \\
& & GraLoRA     & & 4e-4 & & & \\
\cmidrule(l){2-8}
& \multirow{2}{*}{LLaMA3.1–70B} & LoRA & \multirow{2}{*}{64} & \multirow{2}{*}{3e-4} & \multirow{2}{*}{192} & \multirow{2}{*}{1} & \multirow{2}{*}{LionW} \\
& & GraLoRA        & & & & & \\
\midrule

\multirow{4}{*}{\makecell{Mathematical \\ Reasoning}} & \multirow{4}{*}{Qwen-2.5-1.5B} & LoRA & \multirow{2}{*}{64} & \multirow{2}{*}{2e-4} & \multirow{2}{*}{192} & \multirow{2}{*}{4} & \multirow{2}{*}{AdamW} \\
& & GraLoRA     & & & & & \\
\cmidrule(l){3-8}
& & LoRA & \multirow{2}{*}{128} & \multirow{2}{*}{4e-4} & \multirow{2}{*}{192} & \multirow{2}{*}{4} & \multirow{2}{*}{AdamW} \\
& & GraLoRA     & & & & & \\
\midrule

\multirow{2}{*}{\makecell{Personalized \\ Image Generation}} & \multirow{2}{*}{SDXL} & LoRA & \multirow{2}{*}{128} & \multirow{2}{*}{1e-4} & \multirow{2}{*}{1} & \multirow{2}{*}{2} & \multirow{2}{*}{AdamW}\\
 & & GraLoRA & & & & & \\

\bottomrule
\end{tabular}
}
\label{tab:general_hyperparams}
\end{table}

\begin{table}[ht]
\centering
\caption{Detailed hyperparameter settings for each sub-tasks in General Language Understanding.}
\resizebox{\textwidth}{!}{
\begin{tabular}{ccccccccccc}
\toprule
\textbf{Model} & \textbf{Task} & \textbf{Method} & \textbf{Rank} & \textbf{LR} & \textbf{Head-LR} & \textbf{Batch size} & \textbf{Epochs} & \textbf{Optimizer}\\
\midrule

\multirow{12}{*}{RoBERTa-base} & \multirow{2}{*}{SST-2} & GraLoRA & \multirow{2}{*}{8} & \multirow{2}{*}{4e-4} & 4e-3 & \multirow{2}{*}{128} & \multirow{2}{*}{60} & \multirow{2}{*}{AdamW}\\
 & & Hybrid GraLoRA & & & 4e-4 & & \\
\cmidrule(l){2-9}

& \multirow{2}{*}{MRPC}  & GraLoRA & \multirow{2}{*}{8} & \multirow{2}{*}{4e-4} & 4e-4 & \multirow{2}{*}{128} & \multirow{2}{*}{30} & \multirow{2}{*}{AdamW}\\
& & Hybrid GraLoRA & & & 2e-4 & & & \\
\cmidrule(l){2-9}

& \multirow{2}{*}{CoLA}  & GraLoRA & \multirow{2}{*}{8} & 5e-4 & 5e-3 & 128 & \multirow{2}{*}{80} & \multirow{2}{*}{AdamW}\\
& & Hybrid GraLoRA & & 8e-4 & 8e-4 & 256 & & \\
\cmidrule(l){2-9}

& \multirow{2}{*}{QNLI}  & GraLoRA & \multirow{2}{*}{8} & \multirow{2}{*}{5e-4} & \multirow{2}{*}{2e-3} & \multirow{2}{*}{128} & \multirow{2}{*}{25} & \multirow{2}{*}{AdamW}\\
& & Hybrid GraLoRA & & & & & & \\
\cmidrule(l){2-9}

& \multirow{2}{*}{RTE}   & GraLoRA & \multirow{2}{*}{8} & \multirow{2}{*}{2e-4} & \multirow{2}{*}{2e-4} & \multirow{2}{*}{128} & \multirow{2}{*}{160} & \multirow{2}{*}{AdamW}\\
& & Hybrid GraLoRA & & & & & & \\
\cmidrule(l){2-9}

& \multirow{2}{*}{STS-B} & GraLoRA & \multirow{2}{*}{8} & \multirow{2}{*}{1e-3} & \multirow{2}{*}{1e-2} & \multirow{2}{*}{128} & \multirow{2}{*}{80} & \multirow{2}{*}{AdamW}\\
& & Hybrid GraLoRA & & & & & & \\
\bottomrule
\end{tabular}
}
\label{tab:glue_hyperparams}
\end{table}

We fixed LoRA \( \alpha=2r \) which is known to be generally applicable in different models with different ranks~\cite{Biderman2024LoRALLFL}. Detailed hyperparameter settings for our experiments are denoted in Table~\ref{tab:general_hyperparams}.

\newpage
\section*{NeurIPS Paper Checklist}

The checklist is designed to encourage best practices for responsible machine learning research, addressing issues of reproducibility, transparency, research ethics, and societal impact. Do not remove the checklist: {\bf The papers not including the checklist will be desk rejected.} The checklist should follow the references and follow the (optional) supplemental material.  The checklist does NOT count towards the page
limit. 

Please read the checklist guidelines carefully for information on how to answer these questions. For each question in the checklist:
\begin{itemize}
    \item You should answer \answerYes{}, \answerNo{}, or \answerNA{}.
    \item \answerNA{} means either that the question is Not Applicable for that particular paper or the relevant information is Not Available.
    \item Please provide a short (1–2 sentence) justification right after your answer (even for NA). 
\end{itemize}

{\bf The checklist answers are an integral part of your paper submission.} They are visible to the reviewers, area chairs, senior area chairs, and ethics reviewers. You will be asked to also include it (after eventual revisions) with the final version of your paper, and its final version will be published with the paper.

The reviewers of your paper will be asked to use the checklist as one of the factors in their evaluation. While "\answerYes{}" is generally preferable to "\answerNo{}", it is perfectly acceptable to answer "\answerNo{}" provided a proper justification is given (e.g., "error bars are not reported because it would be too computationally expensive" or "we were unable to find the license for the dataset we used"). In general, answering "\answerNo{}" or "\answerNA{}" is not grounds for rejection. While the questions are phrased in a binary way, we acknowledge that the true answer is often more nuanced, so please just use your best judgment and write a justification to elaborate. All supporting evidence can appear either in the main paper or the supplemental material, provided in appendix. If you answer \answerYes{} to a question, in the justification please point to the section(s) where related material for the question can be found.

IMPORTANT, please:
\begin{itemize}
    \item {\bf Delete this instruction block, but keep the section heading ``NeurIPS Paper Checklist"},
    \item  {\bf Keep the checklist subsection headings, questions/answers and guidelines below.}
    \item {\bf Do not modify the questions and only use the provided macros for your answers}.
\end{itemize}


\begin{enumerate}

\item {\bf Claims}
    \item[] Question: Do the main claims made in the abstract and introduction accurately reflect the paper's contributions and scope?
    \item[] Answer: \answerYes{} 
    \item[] Justification: Yes, we strongly denote our main claims in the abstract and introduction. 
    \item[] Guidelines:
    \begin{itemize}
        \item The answer NA means that the abstract and introduction do not include the claims made in the paper.
        \item The abstract and/or introduction should clearly state the claims made, including the contributions made in the paper and important assumptions and limitations. A No or NA answer to this question will not be perceived well by the reviewers. 
        \item The claims made should match theoretical and experimental results, and reflect how much the results can be expected to generalize to other settings. 
        \item It is fine to include aspirational goals as motivation as long as it is clear that these goals are not attained by the paper. 
    \end{itemize}

\item {\bf Limitations}
    \item[] Question: Does the paper discuss the limitations of the work performed by the authors?
    \item[] Answer: \answerYes{} 
    \item[] Justification: Yes, we discuss the limitations of our method and evaluate it's impact in the "Tradeoff Analysis" section. We further examine how to overcome the limitation in "Hybrid GraLoRA" section and it's practical result in "Results on Code Generation" section.
    \item[] Guidelines:
    \begin{itemize}
        \item The answer NA means that the paper has no limitation while the answer No means that the paper has limitations, but those are not discussed in the paper. 
        \item The authors are encouraged to create a separate "Limitations" section in their paper.
        \item The paper should point out any strong assumptions and how robust the results are to violations of these assumptions (e.g., independence assumptions, noiseless settings, model well-specification, asymptotic approximations only holding locally). The authors should reflect on how these assumptions might be violated in practice and what the implications would be.
        \item The authors should reflect on the scope of the claims made, e.g., if the approach was only tested on a few datasets or with a few runs. In general, empirical results often depend on implicit assumptions, which should be articulated.
        \item The authors should reflect on the factors that influence the performance of the approach. For example, a facial recognition algorithm may perform poorly when image resolution is low or images are taken in low lighting. Or a speech-to-text system might not be used reliably to provide closed captions for online lectures because it fails to handle technical jargon.
        \item The authors should discuss the computational efficiency of the proposed algorithms and how they scale with dataset size.
        \item If applicable, the authors should discuss possible limitations of their approach to address problems of privacy and fairness.
        \item While the authors might fear that complete honesty about limitations might be used by reviewers as grounds for rejection, a worse outcome might be that reviewers discover limitations that aren't acknowledged in the paper. The authors should use their best judgment and recognize that individual actions in favor of transparency play an important role in developing norms that preserve the integrity of the community. Reviewers will be specifically instructed to not penalize honesty concerning limitations.
    \end{itemize}

\item {\bf Theory assumptions and proofs}
    \item[] Question: For each theoretical result, does the paper provide the full set of assumptions and a complete (and correct) proof?
    \item[] Answer: \answerYes{} 
    \item[] Justification: Yes, we do provide complete and correct proof for each theoretical result in the "Expression Power Analysis" section.
    \item[] Guidelines:
    \begin{itemize}
        \item The answer NA means that the paper does not include theoretical results. 
        \item All the theorems, formulas, and proofs in the paper should be numbered and cross-referenced.
        \item All assumptions should be clearly stated or referenced in the statement of any theorems.
        \item The proofs can either appear in the main paper or the supplemental material, but if they appear in the supplemental material, the authors are encouraged to provide a short proof sketch to provide intuition. 
        \item Inversely, any informal proof provided in the core of the paper should be complemented by formal proofs provided in appendix or supplemental material.
        \item Theorems and Lemmas that the proof relies upon should be properly referenced. 
    \end{itemize}

    \item {\bf Experimental result reproducibility}
    \item[] Question: Does the paper fully disclose all the information needed to reproduce the main experimental results of the paper to the extent that it affects the main claims and/or conclusions of the paper (regardless of whether the code and data are provided or not)?
    \item[] Answer: \answerYes{} 
    \item[] Justification: Yes, we do disclose information needed to reproduce the results in "Experiment Setup" section and in Appendix.
    \item[] Guidelines:
    \begin{itemize}
        \item The answer NA means that the paper does not include experiments.
        \item If the paper includes experiments, a No answer to this question will not be perceived well by the reviewers: Making the paper reproducible is important, regardless of whether the code and data are provided or not.
        \item If the contribution is a dataset and/or model, the authors should describe the steps taken to make their results reproducible or verifiable. 
        \item Depending on the contribution, reproducibility can be accomplished in various ways. For example, if the contribution is a novel architecture, describing the architecture fully might suffice, or if the contribution is a specific model and empirical evaluation, it may be necessary to either make it possible for others to replicate the model with the same dataset, or provide access to the model. In general. releasing code and data is often one good way to accomplish this, but reproducibility can also be provided via detailed instructions for how to replicate the results, access to a hosted model (e.g., in the case of a large language model), releasing of a model checkpoint, or other means that are appropriate to the research performed.
        \item While NeurIPS does not require releasing code, the conference does require all submissions to provide some reasonable avenue for reproducibility, which may depend on the nature of the contribution. For example
        \begin{enumerate}
            \item If the contribution is primarily a new algorithm, the paper should make it clear how to reproduce that algorithm.
            \item If the contribution is primarily a new model architecture, the paper should describe the architecture clearly and fully.
            \item If the contribution is a new model (e.g., a large language model), then there should either be a way to access this model for reproducing the results or a way to reproduce the model (e.g., with an open-source dataset or instructions for how to construct the dataset).
            \item We recognize that reproducibility may be tricky in some cases, in which case authors are welcome to describe the particular way they provide for reproducibility. In the case of closed-source models, it may be that access to the model is limited in some way (e.g., to registered users), but it should be possible for other researchers to have some path to reproducing or verifying the results.
        \end{enumerate}
    \end{itemize}

\item {\bf Open access to data and code}
    \item[] Question: Does the paper provide open access to the data and code, with sufficient instructions to faithfully reproduce the main experimental results, as described in supplemental material?
    \item[] Answer: \answerYes{} 
    \item[] Justification: Yes, we do provide open access to code with sufficient instructions as supplemental material.
    \item[] Guidelines:
    \begin{itemize}
        \item The answer NA means that paper does not include experiments requiring code.
        \item Please see the NeurIPS code and data submission guidelines (\url{https://nips.cc/public/guides/CodeSubmissionPolicy}) for more details.
        \item While we encourage the release of code and data, we understand that this might not be possible, so “No” is an acceptable answer. Papers cannot be rejected simply for not including code, unless this is central to the contribution (e.g., for a new open-source benchmark).
        \item The instructions should contain the exact command and environment needed to run to reproduce the results. See the NeurIPS code and data submission guidelines (\url{https://nips.cc/public/guides/CodeSubmissionPolicy}) for more details.
        \item The authors should provide instructions on data access and preparation, including how to access the raw data, preprocessed data, intermediate data, and generated data, etc.
        \item The authors should provide scripts to reproduce all experimental results for the new proposed method and baselines. If only a subset of experiments are reproducible, they should state which ones are omitted from the script and why.
        \item At submission time, to preserve anonymity, the authors should release anonymized versions (if applicable).
        \item Providing as much information as possible in supplemental material (appended to the paper) is recommended, but including URLs to data and code is permitted.
    \end{itemize}

\item {\bf Experimental setting/details}
    \item[] Question: Does the paper specify all the training and test details (e.g., data splits, hyperparameters, how they were chosen, type of optimizer, etc.) necessary to understand the results?
    \item[] Answer: \answerYes{} 
    \item[] Justification: Yes, we specify all the training and test details necessary to understand the results in "Experimental Setup" section and in Appendix.
    \item[] Guidelines:
    \begin{itemize}
        \item The answer NA means that the paper does not include experiments.
        \item The experimental setting should be presented in the core of the paper to a level of detail that is necessary to appreciate the results and make sense of them.
        \item The full details can be provided either with the code, in appendix, or as supplemental material.
    \end{itemize}

\item {\bf Experiment statistical significance}
    \item[] Question: Does the paper report error bars suitably and correctly defined or other appropriate information about the statistical significance of the experiments?
    \item[] Answer: \answerNo{} 
    \item[] Justification: No, error bars are not reported because it would be too computationally expensive. 
    \item[] Guidelines:
    \begin{itemize}
        \item The answer NA means that the paper does not include experiments.
        \item The authors should answer "Yes" if the results are accompanied by error bars, confidence intervals, or statistical significance tests, at least for the experiments that support the main claims of the paper.
        \item The factors of variability that the error bars are capturing should be clearly stated (for example, train/test split, initialization, random drawing of some parameter, or overall run with given experimental conditions).
        \item The method for calculating the error bars should be explained (closed form formula, call to a library function, bootstrap, etc.)
        \item The assumptions made should be given (e.g., Normally distributed errors).
        \item It should be clear whether the error bar is the standard deviation or the standard error of the mean.
        \item It is OK to report 1-sigma error bars, but one should state it. The authors should preferably report a 2-sigma error bar than state that they have a 96\% CI, if the hypothesis of Normality of errors is not verified.
        \item For asymmetric distributions, the authors should be careful not to show in tables or figures symmetric error bars that would yield results that are out of range (e.g. negative error rates).
        \item If error bars are reported in tables or plots, The authors should explain in the text how they were calculated and reference the corresponding figures or tables in the text.
    \end{itemize}

\item {\bf Experiments compute resources}
    \item[] Question: For each experiment, does the paper provide sufficient information on the computer resources (type of compute workers, memory, time of execution) needed to reproduce the experiments?
    \item[] Answer: \answerYes{} 
    \item[] Justification: Yes, we do provide sufficient information on the computer resources in "Tradeoff Analysis" section and in "Experiments" section.
    \item[] Guidelines:
    \begin{itemize}
        \item The answer NA means that the paper does not include experiments.
        \item The paper should indicate the type of compute workers CPU or GPU, internal cluster, or cloud provider, including relevant memory and storage.
        \item The paper should provide the amount of compute required for each of the individual experimental runs as well as estimate the total compute. 
        \item The paper should disclose whether the full research project required more compute than the experiments reported in the paper (e.g., preliminary or failed experiments that didn't make it into the paper). 
    \end{itemize}
    
\item {\bf Code of ethics}
    \item[] Question: Does the research conducted in the paper conform, in every respect, with the NeurIPS Code of Ethics \url{https://neurips.cc/public/EthicsGuidelines}?
    \item[] Answer: \answerYes{} 
    \item[] Justification: Yes, our research conforms with the NeurIPS Code of Ethics.
    \item[] Guidelines:
    \begin{itemize}
        \item The answer NA means that the authors have not reviewed the NeurIPS Code of Ethics.
        \item If the authors answer No, they should explain the special circumstances that require a deviation from the Code of Ethics.
        \item The authors should make sure to preserve anonymity (e.g., if there is a special consideration due to laws or regulations in their jurisdiction).
    \end{itemize}

\item {\bf Broader impacts}
    \item[] Question: Does the paper discuss both potential positive societal impacts and negative societal impacts of the work performed?
    \item[] Answer: \answerNA{} 
    \item[] Justification: We propose a generic method for optimizing efficient training of neural networks. It is not directly related to social impacts.
    \item[] Guidelines:
    \begin{itemize}
        \item The answer NA means that there is no societal impact of the work performed.
        \item If the authors answer NA or No, they should explain why their work has no societal impact or why the paper does not address societal impact.
        \item Examples of negative societal impacts include potential malicious or unintended uses (e.g., disinformation, generating fake profiles, surveillance), fairness considerations (e.g., deployment of technologies that could make decisions that unfairly impact specific groups), privacy considerations, and security considerations.
        \item The conference expects that many papers will be foundational research and not tied to particular applications, let alone deployments. However, if there is a direct path to any negative applications, the authors should point it out. For example, it is legitimate to point out that an improvement in the quality of generative models could be used to generate deepfakes for disinformation. On the other hand, it is not needed to point out that a generic algorithm for optimizing neural networks could enable people to train models that generate Deepfakes faster.
        \item The authors should consider possible harms that could arise when the technology is being used as intended and functioning correctly, harms that could arise when the technology is being used as intended but gives incorrect results, and harms following from (intentional or unintentional) misuse of the technology.
        \item If there are negative societal impacts, the authors could also discuss possible mitigation strategies (e.g., gated release of models, providing defenses in addition to attacks, mechanisms for monitoring misuse, mechanisms to monitor how a system learns from feedback over time, improving the efficiency and accessibility of ML).
    \end{itemize}
    
\item {\bf Safeguards}
    \item[] Question: Does the paper describe safeguards that have been put in place for responsible release of data or models that have a high risk for misuse (e.g., pretrained language models, image generators, or scraped datasets)?
    \item[] Answer: \answerNA{} 
    \item[] Justification: Our method solely works for fine-tuning a pretrained model, thus the paper poses no such risks.
    \item[] Guidelines:
    \begin{itemize}
        \item The answer NA means that the paper poses no such risks.
        \item Released models that have a high risk for misuse or dual-use should be released with necessary safeguards to allow for controlled use of the model, for example by requiring that users adhere to usage guidelines or restrictions to access the model or implementing safety filters. 
        \item Datasets that have been scraped from the Internet could pose safety risks. The authors should describe how they avoided releasing unsafe images.
        \item We recognize that providing effective safeguards is challenging, and many papers do not require this, but we encourage authors to take this into account and make a best faith effort.
    \end{itemize}

\item {\bf Licenses for existing assets}
    \item[] Question: Are the creators or original owners of assets (e.g., code, data, models), used in the paper, properly credited and are the license and terms of use explicitly mentioned and properly respected?
    \item[] Answer: \answerYes{} 
    \item[] Justification: Yes, the assets used in the paper are properly cited in "Experiments" section and in overall writings.
    \item[] Guidelines:
    \begin{itemize}
        \item The answer NA means that the paper does not use existing assets.
        \item The authors should cite the original paper that produced the code package or dataset.
        \item The authors should state which version of the asset is used and, if possible, include a URL.
        \item The name of the license (e.g., CC-BY 4.0) should be included for each asset.
        \item For scraped data from a particular source (e.g., website), the copyright and terms of service of that source should be provided.
        \item If assets are released, the license, copyright information, and terms of use in the package should be provided. For popular datasets, \url{paperswithcode.com/datasets} has curated licenses for some datasets. Their licensing guide can help determine the license of a dataset.
        \item For existing datasets that are re-packaged, both the original license and the license of the derived asset (if it has changed) should be provided.
        \item If this information is not available online, the authors are encouraged to reach out to the asset's creators.
    \end{itemize}

\item {\bf New assets}
    \item[] Question: Are new assets introduced in the paper well documented and is the documentation provided alongside the assets?
    \item[] Answer: \answerNA{} 
    \item[] Justification: We do not introduce new assets.
    \item[] Guidelines:
    \begin{itemize}
        \item The answer NA means that the paper does not release new assets.
        \item Researchers should communicate the details of the dataset/code/model as part of their submissions via structured templates. This includes details about training, license, limitations, etc. 
        \item The paper should discuss whether and how consent was obtained from people whose asset is used.
        \item At submission time, remember to anonymize your assets (if applicable). You can either create an anonymized URL or include an anonymized zip file.
    \end{itemize}

\item {\bf Crowdsourcing and research with human subjects}
    \item[] Question: For crowdsourcing experiments and research with human subjects, does the paper include the full text of instructions given to participants and screenshots, if applicable, as well as details about compensation (if any)? 
    \item[] Answer: \answerNA{} 
    \item[] Justification: We do not involve crowdsourcing nor research with human subjects.
    \item[] Guidelines:
    \begin{itemize}
        \item The answer NA means that the paper does not involve crowdsourcing nor research with human subjects.
        \item Including this information in the supplemental material is fine, but if the main contribution of the paper involves human subjects, then as much detail as possible should be included in the main paper. 
        \item According to the NeurIPS Code of Ethics, workers involved in data collection, curation, or other labor should be paid at least the minimum wage in the country of the data collector. 
    \end{itemize}

\item {\bf Institutional review board (IRB) approvals or equivalent for research with human subjects}
    \item[] Question: Does the paper describe potential risks incurred by study participants, whether such risks were disclosed to the subjects, and whether Institutional Review Board (IRB) approvals (or an equivalent approval/review based on the requirements of your country or institution) were obtained?
    \item[] Answer: \answerNA{} 
    \item[] Justification: We do not involve crowdsourcing nor research with human subjects.
    \item[] Guidelines:
    \begin{itemize}
        \item The answer NA means that the paper does not involve crowdsourcing nor research with human subjects.
        \item Depending on the country in which research is conducted, IRB approval (or equivalent) may be required for any human subjects research. If you obtained IRB approval, you should clearly state this in the paper. 
        \item We recognize that the procedures for this may vary significantly between institutions and locations, and we expect authors to adhere to the NeurIPS Code of Ethics and the guidelines for their institution. 
        \item For initial submissions, do not include any information that would break anonymity (if applicable), such as the institution conducting the review.
    \end{itemize}

\item {\bf Declaration of LLM usage}
    \item[] Question: Does the paper describe the usage of LLMs if it is an important, original, or non-standard component of the core methods in this research? Note that if the LLM is used only for writing, editing, or formatting purposes and does not impact the core methodology, scientific rigorousness, or originality of the research, declaration is not required.
    \item[] Answer: \answerNA{} 
    \item[] Justification: Our core method development does not involve LLMs.
    \item[] Guidelines:
    \begin{itemize}
        \item The answer NA means that the core method development in this research does not involve LLMs as any important, original, or non-standard components.
        \item Please refer to our LLM policy (\url{https://neurips.cc/Conferences/2025/LLM}) for what should or should not be described.
    \end{itemize}

\end{enumerate}

\end{document}